\documentclass[11pt]{article}

\usepackage[preprint]{acl}

\usepackage{times}
\usepackage{latexsym}
\usepackage[T1]{fontenc}
\usepackage[utf8]{inputenc}
\usepackage{microtype}
\usepackage{inconsolata}

\usepackage{graphicx}
\usepackage{booktabs}
\usepackage{multirow}
\usepackage{amsfonts}
\usepackage{amsmath}
\usepackage{nicefrac}
\usepackage{xcolor}
\usepackage{url}
\usepackage{enumitem}
\usepackage{eso-pic}

\newcommand{\caseimage}[2][0.42\linewidth]{%
\begin{center}
\includegraphics[width=#1,height=0.9in,keepaspectratio]{#2}
\end{center}
}

\definecolor{PromptBg}{HTML}{F3F4F6}
\definecolor{PromptBorder}{HTML}{9CA3AF}
\definecolor{ReferenceBg}{HTML}{E8F1FB}
\definecolor{ReferenceBorder}{HTML}{3B82C4}
\definecolor{StudentBg}{HTML}{FFF3DC}
\definecolor{StudentBorder}{HTML}{D99022}
\definecolor{RecoBg}{HTML}{E8F6EE}
\definecolor{RecoBorder}{HTML}{2F8F5B}
\definecolor{AnchorBg}{HTML}{FDECEC}
\definecolor{AnchorBorder}{HTML}{C94A4A}
\newcommand{\casebox}[4]{%
\noindent\fcolorbox{#1}{#2}{%
\begin{minipage}{0.94\linewidth}
\scriptsize \textbf{#3}\par\vspace{1pt}#4
\end{minipage}}%
}
\newcommand{\qualbox}[5]{%
\noindent\fcolorbox{#1}{#2}{%
\begin{minipage}{#3}
\scriptsize\raggedright \textbf{#4}\par\vspace{1pt}#5
\end{minipage}}%
}
\newcommand{\authoraffil}[1]{{\normalfont\normalsize #1}}
\newcommand{\authorcorr}[1]{{\normalfont\small #1}}

\title{\textbf{RoCo-ACE}: Rollout-Conditioned Online Distillation for Retention-Aware Knowledge Injection}
\author{
{\bfseries Yan Hong$^{1}$, Wei Li$^{1}$, Kedong Xiu$^{2}$, Jun Lan$^{*,1}$, Shuheng Zhou$^{1}$}\\
{\bfseries Zhongcai Lyu$^{1}$, Huijia Zhu$^{1}$, Weiqiang Wang$^{1}$, Jianfu Zhang$^{*,3}$}\\[0.35em]
\authoraffil{$^{1}$Ant Group}\\
\authoraffil{$^{2}$Zhejiang University}\\
\authoraffil{$^{3}$Shanghai Jiao Tong University}\\[0.15em]
\authorcorr{\textbf{Correspondence:} Jun Lan \href{mailto:yelan.lj@antgroup.com}{\textless yelan.lj@antgroup.com\textgreater}, Jianfu Zhang \href{mailto:c.sis@sjtu.edu.cn}{\textless c.sis@sjtu.edu.cn\textgreater}}
}

\begin{document}
\maketitle
\AddToShipoutPictureFG*{%
  \AtPageLowerLeft{%
    \raisebox{0.62in}{\hspace*{0.78in}\footnotesize\textsuperscript{*}Corresponding authors.}%
  }%
}

\begin{abstract}
Knowledge injection updates pretrained MLLMs with new factual or domain-specific knowledge, but fitting full authoritative answers can cause drift in non-updated behavior. Online distillation limits this drift by training on model-generated rollouts, yet uniform reference-conditioned distillation gives coarse supervision: it can under-emphasize reference-supported rollout tokens and supervise omitted facts only indirectly. We introduce \textbf{RoCo-ACE}, a rollout-conditioned online distillation objective for knowledge injection: \textbf{RoCo} uses same-rollout reference-free/reference-conditioned likelihood contrast to reallocate additional distillation weight to reference-supported rollout tokens, while \textbf{ACE} adds sparse reference-side anchored correction for authoritative anchors omitted from the rollout, without full-answer imitation. Across three knowledge-injection settings, six retention benchmarks, multiple baselines and base models, \textbf{RoCo-ACE} achieves the best injected-knowledge accuracy among compared methods while keeping evaluated retention close to the base model.
\end{abstract}


\section{Introduction}
Multimodal Large Language Models (MLLMs) have evolved from vision-language pretraining and image-language bootstrapping~\citep{alayrac2022flamingo,li2023blip2} to instruction-tuned multimodal assistants~\citep{dai2023instructblip,liu2023llava,bai2023qwenvl}.
They now support instruction following, visual understanding, grounding, text-rich image reasoning, and safety-sensitive response behavior~\citep{pi2024mllmprotector,wu2025autosteer}.
After deployment, these models often need to absorb new entities, events, domain facts, and visual-world knowledge.
A useful knowledge-injection update should improve injected-knowledge accuracy while limiting drift in non-updated multimodal and safety behavior, as evaluated by general multimodal benchmarks~\citep{liu2024mibench,chen2024mmstar} and safety benchmarks~\citep{liu2023mmsafetybench}.
This goal is challenging: fitting new data can interfere with previously learned behavior~\citep{kirkpatrick2017overcoming,li2016learning}, and recent MLLM knowledge-injection studies observe the same injection-retention trade-off~\citep{jiang2025mmevoke,jiang2025kore}.

\begin{figure}[t!]
    \centering
    \includegraphics[width=\columnwidth]{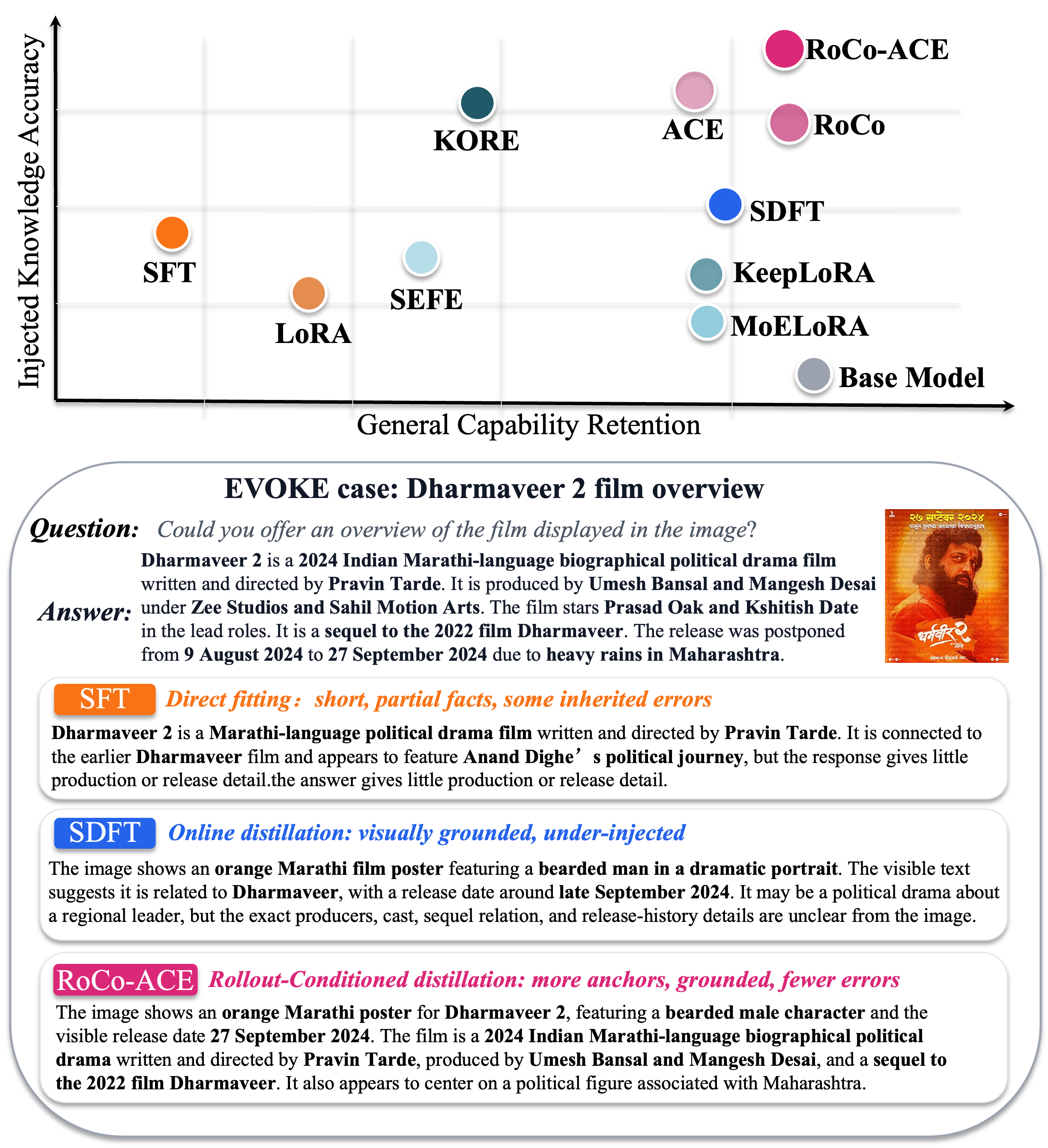}
    \caption{
    Motivation of \textbf{RoCo-ACE}.
    The upper panel illustrates the injection--retention trade-off among direct fitting, constrained updating, online distillation, and our variants.
    \textbf{RoCo-ACE} improves this trade-off by combining same-rollout likelihood contrast with sparse anchored correction.
    The lower panel shows an EVOKE example where \textbf{RoCo-ACE} recovers more authoritative anchors than SFT and SDFT~\citep{shenfeld2026sdft} while keeping a grounded response style.
    }
    \vspace{-16pt}
    \label{fig:motivation}
\end{figure}


Existing update strategies occupy different regions of the injection-retention trade-off.
Supervised Fine-Tuning (SFT)~\citep{ouyang2022training} directly fits authoritative answers and can improve injected-knowledge accuracy, but it also encourages full-answer imitation, including dataset-specific style and narrow response patterns.
Constrained or parameter-efficient methods limit drift by restricting where or how the model can change~\citep{hu2022lora,jiang2025kore,luo2026keeplora}, but they do not directly localize supervision to the factual units that should change.
The key issue is supervision granularity: knowledge injection often requires updating a few entities, dates, relations, or event details, while many training tokens reflect generic phrasing, non-target visual details, or answer style.
As illustrated in Figure~\ref{fig:motivation}, fitting-oriented methods can improve injection at a substantial retention cost, whereas conservative methods better limit drift in non-updated behavior but leave some target facts under-injected.

To address this injection-retention trade-off, we propose \textbf{RoCo-ACE}, a \textbf{Ro}llout-\textbf{Co}nditioned contrastive online distillation objective with \textbf{A}nchored \textbf{C}ross-\textbf{E}ntropy for retention-oriented knowledge injection.
It trains on model-generated rollouts instead of fitting the full authoritative answer, keeping updates close to the current model policy~\citep{agarwal2023gkd,gu2023minillm,shenfeld2026sdft,zhao2026opsd,ye2026opcd}.
However, uniform reference-conditioned online distillation still lacks fine-grained supervision: it can allocate gradient mass to generic or unsupported wording, fail to emphasize reference-supported rollout tokens, and provide only indirect supervision for facts omitted from the current rollout.

Our key observation is that an on-policy rollout exposes a useful supervision split for factual updating.
It may contain reference-supported content already reached on-policy, generic or unsupported wording, and authoritative anchors that are omitted from or weakly matched by the rollout.
\textbf{RoCo} handles the rollout-side part by contrasting reference-free and reference-conditioned teacher views on the same rollout prefix, then reallocating additional distillation weight to tokens whose likelihood increases under the reference-conditioned view.
\textbf{ACE} complements this signal with sparse anchored cross-entropy over extracted authoritative facts, assigning larger weights to omitted or weakly matched anchors.
Together, the two objectives inject sparse factual content without turning the full reference answer into a behavior-cloning target.

This paper makes the following contributions:
\begin{itemize}[leftmargin=*]
    \item We identify supervision granularity as a key issue in online distillation for knowledge injection: uniform rollout distillation can under-emphasize reference-supported tokens and only indirectly train facts omitted from the current rollout.

    \item We introduce \textbf{RoCo-ACE}, a rollout-conditioned contrastive online distillation objective for knowledge injection. \textbf{RoCo} uses same-rollout likelihood contrast to reallocate additional distillation weight to reference-supported rollout tokens, while \textbf{ACE} adds sparse anchored correction for omitted or weakly matched authoritative facts.

    \item Experiments on three knowledge-injection settings and six retention benchmarks show that \textbf{RoCo-ACE} improves injected-knowledge accuracy while keeping evaluated retention close to online-distillation baselines and the base model.
\end{itemize}


\section{Related Work}




\subsection{Update-Space Constraints for Retention}
A direct way to inject knowledge is to fit authoritative answers, but full-answer fitting can shift non-target behavior.
To limit such update-induced drift, prior work often constrains where or how the model can change.
Adam-NSCL~\citep{wang2021nscl} updates networks in the null space of feature covariance, and LoRA-Null~\citep{tang2025loranull} applies a related idea to low-rank adaptation by choosing initialization spaces that better retain pretrained behavior.
KORE~\citep{jiang2025kore} studies multimodal knowledge injection directly and combines knowledge-oriented augmentation with constraints for limiting drift in multimodal updates.
Parameter-efficient and forgetting-aware methods, including LoRA~\citep{hu2022lora}, KeepLoRA~\citep{luo2026keeplora}, MoELoRA~\citep{luo2024moelora}, and SEFE~\citep{chen2025sefe}, restrict or organize trainable parameters through low-rank adapters, residual-gradient updates, expertized modules, or forgetting-aware constraints.
Related MLLM continual-learning work, such as MLLM-CL~\citep{zhao2025mllmcl} and LiLoRA~\citep{che2025lilora}, further studies sequential visual-instruction updates over heterogeneous task streams.

These methods are retention-oriented mainly through constraints on the editable parameter or update space.
\textbf{RoCo-ACE} is complementary: rather than constraining the update space, it improves supervision granularity over rollout tokens and authoritative anchors.

\subsection{Online Distillation for Model Updating}
Knowledge distillation~\citep{hinton2015distilling} transfers a teacher distribution to a student.
Online or on-policy variants, such as Generalized Knowledge Distillation~\citep{agarwal2023gkd} and MiniLLM~\citep{gu2023minillm}, distill on student-generated outputs rather than only fixed target responses, keeping supervision close to the current student policy.
Recent work further studies richer teacher contexts, reasoning compression, and policy-optimization variants~\citep{zhao2026opsd,ye2026opcd,sang2026crisp,yang2026rlsd,li2026srpo}. \citet{song2026opdsurvey} provides a recent survey.
SDFT~\citep{shenfeld2026sdft} is especially related, as it uses self-distillation to stabilize continual updates and reduce forgetting.

Online distillation is attractive for knowledge injection because it avoids directly fitting the full authoritative answer.
However, uniform reference-conditioned rollout distillation still provides coarse supervision: it can spend gradients on generic or unsupported wording, under-emphasize reference-supported rollout tokens, and supervise omitted authoritative facts only indirectly.
\textbf{RoCo-ACE} improves this granularity by combining same-rollout likelihood contrast for reference-supported rollout tokens with sparse reference-side correction for authoritative anchors omitted from or weakly matched by the rollout.


\begin{figure*}[t]
    \centering
    \includegraphics[width=\textwidth]{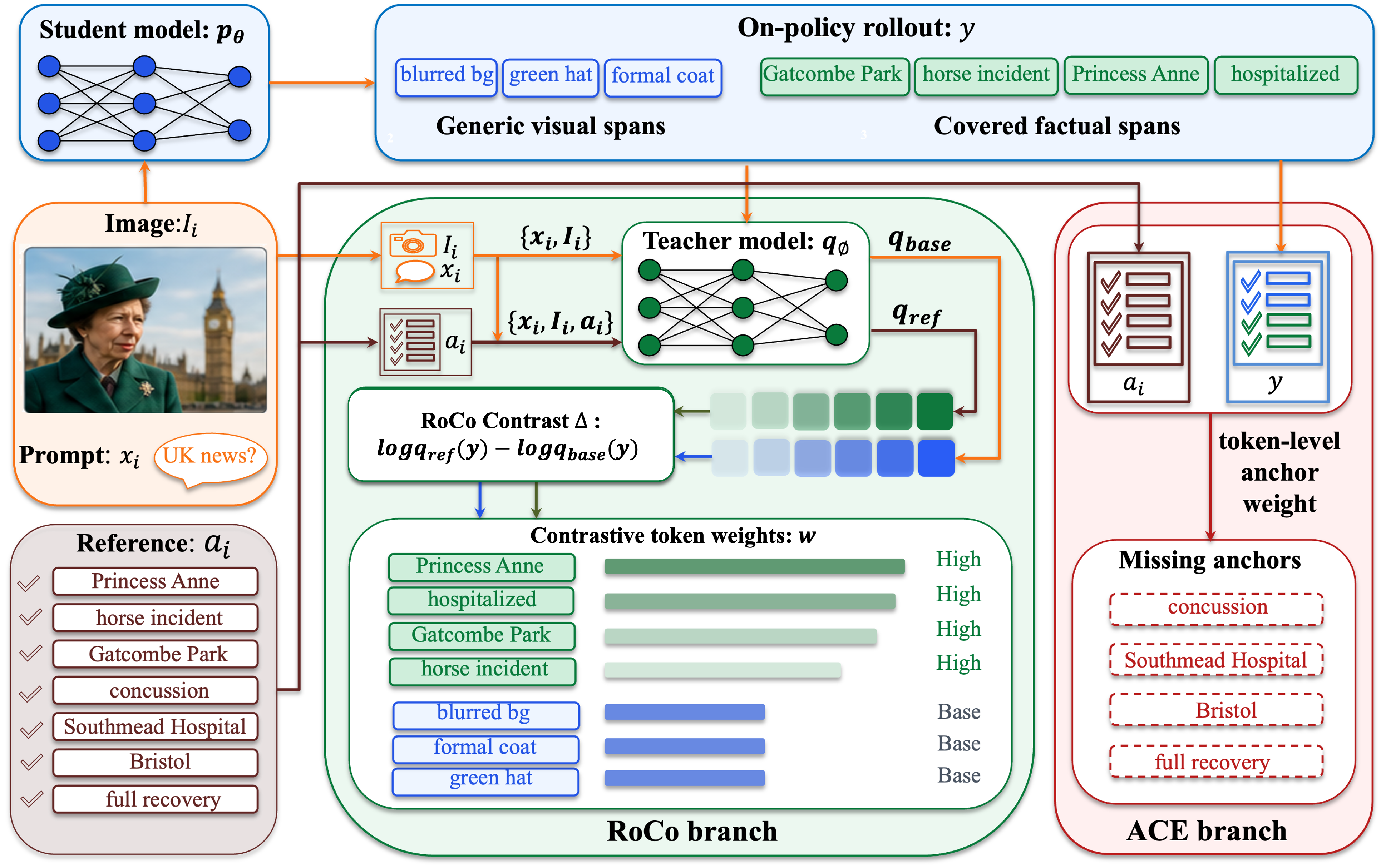}
    \vspace{-16pt}
    \caption{
    Overview of \textbf{RoCo-ACE}.
    The student samples an on-policy rollout that mixes generic or non-target visual spans, reference-supported factual spans, and omitted authoritative anchors.
    \textbf{RoCo} scores the same rollout under reference-free and reference-conditioned teacher views, then uses their likelihood contrast to assign larger distillation weights to reference-supported rollout tokens.
    \textbf{ACE} complements this rollout-side signal with sparse anchored cross-entropy for authoritative anchors omitted from or weakly matched by the rollout.
    }
    \vspace{-8pt}
    \label{fig:method_overview}
\end{figure*}



\section{Method}

\subsection{Overview}

We consider a knowledge-injection dataset $\mathcal{D}=\{(x_i,a_i,I_i)\}_{i=1}^{N}$, where $x_i$ is a user prompt, $a_i$ is an authoritative reference answer that specifies the target factual update, and $I_i$ is an optional image.
Our goal is to adapt a base MLLM $p_{\theta_0}$ into a student model $p_\theta$ that incorporates the injected facts into its responses while limiting drift in non-updated behavior.
Figure~\ref{fig:method_overview} gives an overview of the \textbf{RoCo-ACE} training objective.
For each example, the student samples an on-policy rollout $y_i=(y_{i,1},\ldots,y_{i,T_i}) \sim p_\theta(\cdot \mid x_i,I_i)$.
This rollout exposes a supervision split for factual updating: it may contain reference-supported factual spans already reached on-policy, generic or unsupported wording and non-target visual details, and authoritative anchors that are omitted from or weakly matched by the rollout.
\textbf{RoCo} handles the rollout-side signal by scoring the same rollout prefixes with reference-free and reference-conditioned teacher views.
It uses their likelihood contrast to reallocate additional distillation weight to reference-supported rollout tokens, while generic wording and non-target visual details receive mainly the floor distillation signal.
\textbf{ACE} handles the omission case with sparse anchored cross-entropy over extracted authoritative facts, assigning larger weights to anchors omitted from or weakly matched by the rollout.
Together, \textbf{RoCo} and \textbf{ACE} inject sparse factual content without treating the full reference answer as a token-level imitation target.

\subsection{RoCo: Rollout-Conditioned Contrastive Online Distillation}
Online distillation~\citep{shenfeld2026sdft,zhao2026opsd} is well suited to retention-oriented knowledge injection because it trains on the student's own rollout instead of fitting the full authoritative answer.
This keeps the update close to the current student policy, but uniform reference-conditioned distillation still assigns loss to every valid rollout token, even though only a small subset carries target facts.
A student rollout often mixes reference-supported factual content with generic or unsupported wording.
In Figure~\ref{fig:method_overview}, for example, the rollout contains reference-supported spans such as ``Princess Anne'', ``hospitalized'', and ``Gatcombe Park'', but also includes non-target visual details such as ``dark green hat'' and ``formal coat''.
\textbf{RoCo} addresses this supervision-granularity problem by comparing reference-free and reference-conditioned teacher likelihoods under the same rollout prefix.
It reallocates additional distillation weight to tokens whose likelihood increases when the authoritative reference is available, while generic wording, non-target visual details, and answer style receive mainly the floor distillation signal.
To make this selective reallocation explicit, \textbf{RoCo} measures how much the authoritative reference changes the likelihood assigned to each on-policy token under the same rollout prefix.
Let $q_\phi$ denote the teacher model, and let $q_\phi(\cdot \mid \cdot)$ denote its next-token probability distribution over the vocabulary.
For each rollout prefix $y_{i,<t}$, we evaluate two matched teacher views:
\begin{equation}
\begin{aligned}
q_{\mathrm{base}}^{i,t}(v)
&= q_\phi(v \mid x_i,I_i,y_{i,<t}),\\
q_{\mathrm{ref}}^{i,t}(v)
&= q_\phi(v \mid x_i,I_i,a_i,y_{i,<t}).
\end{aligned}
\end{equation}
The subscript $\mathrm{base}$ denotes the reference-free view, and $\mathrm{ref}$ denotes the reference-conditioned view.
Both views score the same student-generated prefix and share the same architecture and parameters. They differ only in whether the authoritative reference $a_i$ is included in the prompt context.
In our online setting, $q_\phi$ is an exponential-moving-average teacher synchronized from the student, keeping teacher scoring close to the current policy while avoiding a degenerate self-target at the current optimization step.
For each generated token $y_{i,t}$, the RoCo contrast is defined as the log-likelihood ratio between the two views:
\begin{equation}
\Delta_{i,t}
= \log q_{\mathrm{ref}}^{i,t}(y_{i,t})
- \log q_{\mathrm{base}}^{i,t}(y_{i,t}).
\end{equation}
A high value of $q_{\mathrm{ref}}^{i,t}(y_{i,t})$ alone may simply reflect a generic token that is already likely under the original prompt, whereas a positive $\Delta_{i,t}$ suggests additional support induced by the authoritative reference.
We therefore use the contrast to estimate reference-induced support for rollout tokens: the token is generated on-policy, and the reference-conditioned view makes it more likely under the same teacher and prefix.
Let $\mathcal{V}_i$ denote valid completion-token positions after removing padding and ignored tokens.
The contrast is converted into dense token weights:
\begin{equation}
w_{i,t}
= w_0 + \beta_\Delta \max(0,\Delta_{i,t}-\tau),
\quad t\in\mathcal{V}_i,
\end{equation}
where $w_0$ is a floor weight, $\beta_\Delta$ scales the contrast signal, and $\tau$ suppresses small log-likelihood differences.
The floor term keeps a stable online-distillation signal over the rollout, while the contrast term reallocates additional gradient mass to tokens whose likelihood becomes sufficiently larger under the reference-conditioned view.
When a detected factual expression is split into multiple subword tokens, we aggregate the token weights within the expression and broadcast the resulting span score back to its tokens.
For simplicity, we still denote the resulting weights by $w_{i,t}$.
This span-aware smoothing reduces subword-level noise for names, dates, numbers, and multilingual expressions.

The resulting distillation objective is
\begin{equation}
\begin{aligned}
\mathcal{L}_{\mathrm{RoCo}}^{(i)}
&= \frac{\sum_{t\in\mathcal{V}_i} w_{i,t} d_{i,t}}
{\sum_{t\in\mathcal{V}_i} w_{i,t}+\epsilon}, \\
d_{i,t}
&= D_{\mathrm{KL}}\!\left(
q_{\mathrm{ref}}^{i,t}(\cdot)
\Vert
p_\theta(\cdot \mid x_i,I_i,y_{i,<t})
\right).
\end{aligned}
\end{equation}
The denominator makes the objective a weighted average over valid rollout tokens, so RoCo changes the relative distribution of distillation weight rather than simply changing the overall loss scale.
The reference-conditioned teacher provides the distributional target, while the reference-versus-base contrast determines where additional distillation weight should concentrate.
Thus, \textbf{RoCo} strengthens reference-supported content already reached on-policy, while keeping training anchored to the student's own completion and down-weighting generic wording, non-target visual details, and answer style.

\subsection{ACE: Anchored Cross-Entropy}
The on-policy design also defines a coverage boundary for \textbf{RoCo}: its loss is evaluated on the valid rollout positions $\mathcal{V}_i$, so its strongest supervision comes from content that the student already reaches in its own completion.
When a target fact is absent from the rollout, reference-conditioned distillation may still influence nearby continuation distributions and help future rollouts move toward that fact, but the current rollout provides only indirect supervision for the omitted anchor.
\textbf{ACE} addresses this remaining coverage gap by applying explicit supervised correction to extracted authoritative anchors, with larger weights for anchors that are missing or weakly covered in the rollout.

Let $a_i=(a_{i,1},\ldots,a_{i,L_i})$ denote the tokenized authoritative answer.
We segment $a_i$ into factual spans $\mathcal{S}_i$.
In our implementation, spans are constructed from entity-like strings, dates, numbers, quoted titles, and compact factual phrases after text normalization. The same normalization is used when comparing reference spans with rollout text.
For each span $s\in\mathcal{S}_i$, let $\rho_i(s,y_i)\in[0,1]$ be its coverage score in the student rollout, where $1$ denotes full coverage and $0$ denotes absence.
We assign each span an anchor weight
\begin{equation}
u_{i,s}
=
u_0 + \beta_{\mathrm{miss}}\big(1-\rho_i(s,y_i)\big),
\end{equation}
where $u_0$ is a base weight for extracted factual anchors and $\beta_{\mathrm{miss}}$ controls the strength of missing- or weak-coverage correction.
Let $\mathcal{I}_i(s)\subseteq\{1,\ldots,L_i\}$ denote the reference-token positions covered by span $s$, and let
$
\mathcal{A}_i=\bigcup_{s\in\mathcal{S}_i}\mathcal{I}_i(s)
$
be the set of anchor-token positions in the reference answer.
We broadcast each span weight to its covered reference tokens. For overlapping spans, the token-level anchor weight is
\begin{equation}
u_{i,k}
=
\max_{s:\,k\in\mathcal{I}_i(s)} u_{i,s},
\quad k\in\mathcal{A}_i .
\end{equation}
The ACE loss for example $i$ is then
\begin{equation}
\mathcal{L}_{\mathrm{ACE}}^{(i)}
=
-\frac{
\sum_{k\in \mathcal{A}_i}
u_{i,k}
\log p_\theta(a_{i,k}\mid x_i,I_i,a_{i,<k})
}{
\sum_{k\in \mathcal{A}_i}u_{i,k}+\epsilon
}.
\end{equation}
Only tokens in $\mathcal{A}_i$ receive supervised loss, and non-anchor reference tokens serve only as teacher-forced context.
Thus, \textbf{ACE} is sparser than ordinary SFT even when $u_0>0$ for extracted anchors.
It complements \textbf{RoCo} by adding reference-side correction for target facts that are absent or weakly covered in the rollout, while \textbf{RoCo} reallocates distillation over student-reachable rollout positions.

\subsection{Overall Objective}
For a mini-batch $\mathcal{B}$, the final training objective combines rollout-side distillation and reference-side anchor correction:
\begin{equation}
\mathcal{L}
=
\frac{1}{|\mathcal{B}|}
\sum_{i\in\mathcal{B}}
\left(
\lambda_{\mathrm{RoCo}}\mathcal{L}_{\mathrm{RoCo}}^{(i)}
+
\lambda_{\mathrm{ACE}}\mathcal{L}_{\mathrm{ACE}}^{(i)}
\right),
\end{equation}
where $\lambda_{\mathrm{RoCo}}\ge0$ and $\lambda_{\mathrm{ACE}}\ge0$ are loss coefficients.
Here, $\mathcal{L}_{\mathrm{RoCo}}^{(i)}$ is evaluated on the valid student-rollout positions $\mathcal{V}_i$, and $\mathcal{L}_{\mathrm{ACE}}^{(i)}$ is evaluated only on the reference anchor-token set $\mathcal{A}_i$.
Together, these two losses provide complementary supervision: \textbf{RoCo} reallocates distillation over reference-aligned reachable rollout positions, while \textbf{ACE} adds sparse reference-side correction for missing or weakly covered anchors, enabling targeted factual injection without full-reference imitation.


\section{Experiments}

\subsection{Settings}

\paragraph{Benchmarks.}

We evaluate knowledge injection along two axes: injected-knowledge acquisition and evaluated retention.
For acquisition, we use EVOKE~\citep{jiang2025mmevoke}, which contains image-grounded entity and news updates with authoritative reference answers, and VP/Sci from MLLM-CL~\citep{zhao2025mllmcl}, which test broader visual-perception and science-domain update settings.
Appendix~\ref{sec:benchmark_details} provides detailed benchmark descriptions.
For retention, we evaluate every trained model on TreeBench~\citep{wang2026treebench}, VStar~\citep{wu2024vstar}, MathVision~\citep{wang2024mathvision}, MMStar~\citep{chen2024mmstar}, BabyVision~\citep{chen2026babyvision}, and MM-SafetyBench~\citep{liu2023mmsafetybench}, abbreviated as MM-Safety in tables.
These benchmarks cover visual grounding, fine-grained perception, multimodal mathematical reasoning, general multimodal understanding, core visual primitives, and multimodal safety behavior; we report their arithmetic mean as Ret. Avg.



\paragraph{Evaluation metrics.}
We report accuracy for injected-knowledge acquisition and evaluated retention.
Open-ended injection responses are judged against the authoritative answer by an LLM judge, which focuses on factual consistency-including entity identity, event identity, date, number, location, and relation.
Benchmark-specific tasks follow their standard answer formats and normalization rules. For multiple-choice tasks, we extract the predicted option from the model response.
Appendix~\ref{sec:evaluation_protocol} provides details of the judge model, answer extraction, and normalization protocol.

\paragraph{Models and baselines.}
The main experiments use \textbf{Qwen3-VL-8B} as the base model; Qwen3-VL-30B results are reported in Appendix~\ref{sec:qwen3vl30b_results}.
We compare direct fitting baselines (LoRA~\citep{hu2022lora} and SFT~\citep{ouyang2022training}), reference-conditioned online distillation (SDFT with authoritative reference~\citep{shenfeld2026sdft}), constrained-update baselines (KORE~\citep{jiang2025kore}, KeepLoRA~\citep{luo2026keeplora}, MoELoRA~\citep{luo2024moelora}, and SEFE~\citep{chen2025sefe}), and our variants, \textbf{RoCo} and \textbf{RoCo-ACE}.
Unless otherwise stated, all methods use the same injected examples, training budget, image preprocessing, and evaluation protocol; only the update rule or supervision objective changes.
KORE is the only exception, as it additionally uses auxiliary general data to estimate its retention constraint.
Implementation details and hyperparameters are provided in Appendix~\ref{sec:training_config}.


\label{sec:main_results}

\begin{table}[!t]
\centering

\scriptsize
\setlength{\tabcolsep}{2.2pt}
\resizebox{\columnwidth}{!}{
\begin{tabular}{lcccccc}
\toprule
\multirow{3}{*}{\textbf{Method}} &
\multicolumn{2}{c}{\textbf{EVOKE}} &
\multicolumn{2}{c}{\textbf{VP}} &
\multicolumn{2}{c}{\textbf{Sci}} \\
\cmidrule(lr){2-3}\cmidrule(lr){4-5}\cmidrule(lr){6-7}
& \textbf{\shortstack{Inj.\\Acc.}} & \textbf{\shortstack{Ret.\\Avg.}}
& \textbf{\shortstack{Inj.\\Acc.}} & \textbf{\shortstack{Ret.\\Avg.}}
& \textbf{\shortstack{Inj.\\Acc.}} & \textbf{\shortstack{Ret.\\Avg.}} \\
\midrule
Qwen3-VL-8B & 16.4 & \textbf{57.2} & 69.9 & \textbf{57.2} & 73.1 & \textbf{57.2} \\
\midrule
LoRA & 18.6 & 45.2 & 74.5 & 54.8 & 79.6 & 52.6 \\
SFT & 20.1 & 31.8 & 72.7 & 51.0 & 83.7 & 51.0 \\
SDFT (w/ ref.) & 20.9 & 55.1 & 74.1 & 56.4 & 81.5 & \textbf{\textit{56.7}} \\
\midrule
KORE & \textbf{\textit{25.2}} & 47.0 & \textbf{\textit{76.4}} & 52.5 & \textbf{\textit{86.7}} & 48.3 \\
KeepLoRA & 17.3 & \textbf{\textit{56.6}} & 74.0 & 52.0 & 74.6 & 54.2 \\
MoELoRA & 17.4 & 48.7 & 71.2 & 55.7 & 74.0 & 55.7 \\
SEFE & 19.8 & 44.8 & 74.8 & 54.3 & 77.8 & 52.1 \\
\midrule
\textbf{RoCo} & 23.7 & \textbf{\textit{56.6}} & 76.2 & \textbf{\textit{56.5}} & 85.6 & 56.4 \\
\textbf{RoCo-ACE} & \textbf{27.6} & 56.5 & \textbf{77.8} & 56.2 & \textbf{87.5} & 56.3 \\
\bottomrule
\end{tabular}
}
\caption{Baseline comparison across three knowledge-injection settings. Inj. Acc. reports injected-knowledge accuracy, and Ret. Avg. averages the six held-out retention benchmarks. Higher is better. Bold and bold-italic mark the best and second-best scores.}
\vspace{-16pt}
\label{tab:main_three_dataset_summary}
\end{table}



\subsection{Baseline Comparison}
\begin{figure}[!t]
    \centering
    \includegraphics[width=\columnwidth]{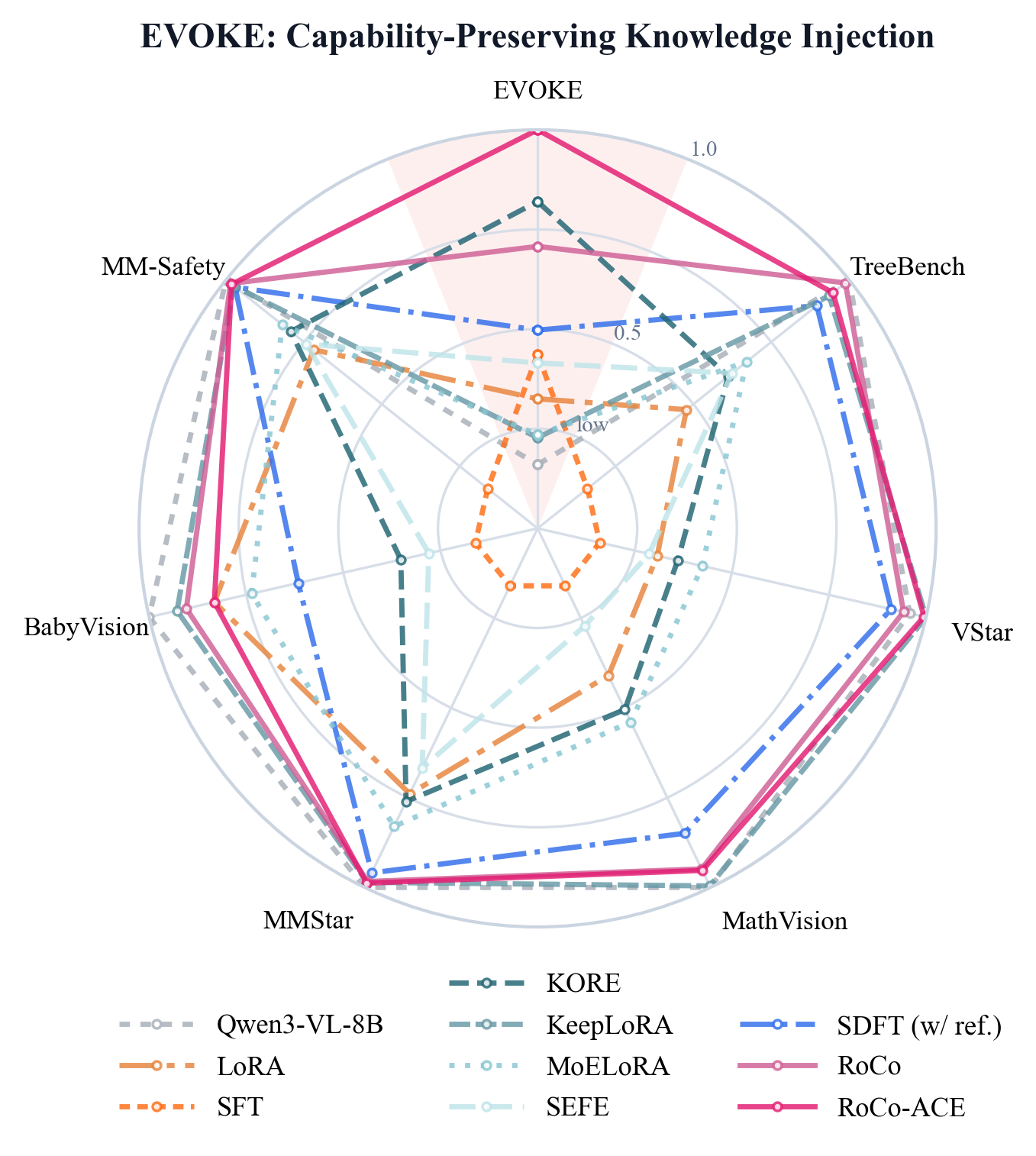}
    \vspace{-16pt}
    \caption{
    Radar comparison on EVOKE.
    The red axis reports injected-knowledge accuracy, and the other axes report evaluated retention on six held-out multimodal and safety benchmarks.
    Scores are normalized for visualization; exact values are reported in Appendix~\ref{app:full_baseline_results}.
    \textbf{RoCo-ACE} improves injected-knowledge accuracy while keeping evaluated retention close to online-distillation and retention-oriented baselines.
    }
    \vspace{-8pt}
    \label{fig:main_baseline_results_evoke}
\end{figure}

Figure~\ref{fig:main_baseline_results_evoke} visualizes the EVOKE injection--retention trade-off, and Table~\ref{tab:main_three_dataset_summary} summarizes Inj. Acc. and Ret. Avg. across EVOKE, VP, and Sci.
Full per-benchmark retention values and VP/Sci radar plots are provided in Appendix~\ref{app:full_baseline_results}. Representative EVOKE responses are shown in Appendix~\ref{sec:baseline_family_examples}.
Overall, Table~\ref{tab:main_three_dataset_summary} shows that existing baselines tend to favor either injection or retention, while \textbf{RoCo-ACE} improves injected-knowledge accuracy with limited drift in evaluated retention.

\noindent\textbf{Fitting-oriented baselines prioritize injection.}
LoRA and SFT directly optimize authoritative answers, making them natural baselines for injected-knowledge acquisition.
However, full-answer supervision also fits generic wording, answer style, and non-target behavior rather than only target factual anchors.
In Table~\ref{tab:main_three_dataset_summary}, SFT improves EVOKE Inj. Acc. from 16.4 to 20.1, but Ret. Avg. drops from 57.2 to 31.8. LoRA has less severe drift, reaching 18.6 Inj. Acc. and 45.2 Ret. Avg.
The same trend appears on VP and Sci, where both direct-fitting baselines improve Inj. Acc. over the base model but reduce Ret. Avg.
These results show that full-answer fitting provides a strong injection signal, but is not ideal for retention-oriented knowledge injection.

\noindent\textbf{Constrained methods can favor stronger injection.}
KORE is the strongest non-ours method for Inj. Acc. in all three settings, reaching 25.2 on EVOKE, 76.4 on VP, and 86.7 on Sci in Table~\ref{tab:main_three_dataset_summary}.
This suggests that knowledge-injection-specific augmentation and constraints can strengthen target acquisition.
The cost is evaluated retention: KORE obtains Ret. Avg. scores of 47.0, 52.5, and 48.3 on EVOKE, VP, and Sci, respectively, lower than the base model and online-distillation baselines.
KORE therefore represents the injection-dominant side of the trade-off.

\noindent\textbf{Retention-oriented baselines limit drift but under-inject.}
SDFT (w/ ref.) and KeepLoRA better maintain Ret. Avg. but are weaker on Inj. Acc.
On EVOKE, SDFT keeps Ret. Avg. at 55.1, close to the base model's 57.2, but reaches only 20.9 Inj. Acc. KeepLoRA reaches 56.6 Ret. Avg. but only 17.3 Inj. Acc.
MoELoRA and SEFE show intermediate behavior across the three settings.
These results are consistent with the limitation of conservative updates or uniform rollout distillation: they can limit evaluated retention drift, but leave some sparse target facts under-injected.

\noindent\textbf{\textbf{RoCo-ACE} improves the trade-off.}
\textbf{RoCo} improves over the SDFT reference point by replacing uniform rollout distillation with same-rollout likelihood contrast.
On EVOKE, it increases Inj. Acc. from 20.9 to 23.7 and Ret. Avg. from 55.1 to 56.6.
Adding \textbf{ACE} further supplies sparse correction for authoritative anchors omitted from or weakly matched by the rollout.
\textbf{RoCo-ACE} achieves the best EVOKE Inj. Acc. among compared methods, 27.6, while keeping Ret. Avg. at 56.5, only 0.7 below the base model.
Compared with KORE, it improves EVOKE Inj. Acc. by 2.4 points and Ret. Avg. by 9.5 points. Compared with SFT, it improves them by 7.5 and 24.7 points.
The same pattern holds on VP and Sci in Table~\ref{tab:main_three_dataset_summary}.
\textbf{RoCo-ACE} obtains the best Inj. Acc. on both settings, 77.8 and 87.5, while keeping Ret. Avg. at 56.2 and 56.3, close to SDFT (w/ ref.) and higher than KORE.
Overall, the results support the central design: \textbf{RoCo} reweights reference-supported rollout tokens through same-rollout likelihood contrast, while \textbf{ACE} adds sparse reference-side correction without full-answer imitation.



\begin{table*}[!t]
\centering

\small
\setlength{\tabcolsep}{4.0pt}
\resizebox{\textwidth}{!}{
\begin{tabular}{lcccccccc}
\toprule
\multirow{2}{*}{\textbf{Variant}} &
\multirow{2}{*}{\textbf{EVOKE}} &
\multicolumn{7}{c}{\textbf{Retention Benchmarks}} \\
\cmidrule(lr){3-9}
& & \textbf{TreeBench} & \textbf{VStar} & \textbf{MathVision} & \textbf{MMStar} & \textbf{BabyVision} & \textbf{MM-Safety} & \textbf{Ret. Avg.} \\
\midrule
Base model & 16.4 & \textbf{43.3} & 83.2 & \textbf{49.3} & \textbf{70.6} & \textbf{\textit{14.2}} & \textbf{\textit{82.6}} & \textbf{57.2} \\
\midrule
SFT & 20.1 & 15.8 & 68.1 & 29.8 & 35.5 & 10.7 & 30.9 & 31.8 \\
SDFT (w/ ref.) & 20.9 & 39.9 & 82.3 & 45.8 & 68.9 & 12.6 & 80.8 & 55.1 \\
SDFT + SFT & 21.8 & 38.1 & 80.6 & 43.2 & 66.1 & 11.9 & 77.9 & 53.0 \\
SDFT + ACE & \textbf{\textit{25.9}} & 40.0 & \textbf{85.1} & 45.2 & 68.8 & 12.5 & 80.3 & 55.3 \\
\textbf{RoCo} & 23.7 & 42.9 & 82.9 & 48.1 & 69.9 & 13.8 & 81.7 & 56.6 \\
\textbf{RoCo-ACE} (mismatched ref.) & 18.5 & 39.5 & 83.1 & 46.3 & 68.9 & 13.5 & 82.0 & 56.8 \\
\textbf{RoCo-ACE} (fixed teacher) & 25.6 & \textbf{\textit{43.0}} & 83.1 & \textbf{\textit{48.9}} & \textbf{\textit{70.1}} & \textbf{14.6} & \textbf{82.8} & \textbf{\textit{57.0}} \\
\textbf{RoCo} + SFT & 24.8 & 41.1 & 82.1 & 45.2 & 65.2 & 12.4 & 80.1 & 54.4 \\
ACE only & 25.1 & 40.6 & 79.5 & 46.8 & 64.6 & 11.9 & 78.5 & 53.6 \\
\textbf{RoCo-ACE} & \textbf{27.6} & 41.6 & \textbf{\textit{83.9}} & 48.2 & \textbf{\textit{70.1}} & 13.5 & 81.5 & 56.5 \\
\bottomrule
\end{tabular}
}
\caption{Ablation results for Qwen3-VL-8B with full retention metrics. EVOKE measures injected-knowledge accuracy. Retention columns report held-out general multimodal and safety accuracy, and Ret. Avg. averages TreeBench, VStar, MathVision, MMStar, BabyVision, and MM-Safety. Higher values are better. Bold and bold-italic mark the best and second-best scores in each column.}
\vspace{-16pt}
\label{tab:ablation_results}
\end{table*}


\subsection{Ablation Study}
Table~\ref{tab:ablation_results} isolates the main design choices.
SDFT (w/ ref.) and \textbf{RoCo} use the same rollout, reference-conditioned teacher target, optimizer, and training data. The key difference is whether rollout tokens receive uniform weights or weights from same-rollout likelihood contrast.
\textbf{RoCo} improves EVOKE Inj. Acc. from 20.9 to 23.7 and Ret. Avg. from 55.1 to 56.6, showing that reweighting reference-supported rollout tokens improves over uniform reference-conditioned distillation.

We further include teacher-side controls.
First, a fixed-teacher variant disables EMA teacher updates while keeping the same \textbf{RoCo-ACE} objective.
Its EVOKE Inj. Acc. drops from 27.6 to 25.6, while Ret. Avg. remains close to the EMA-teacher setting.
This suggests that a fixed teacher can still keep evaluated retention stable, but provides a weaker injection signal than an EMA teacher that tracks the current student during online distillation.

Second, a mismatched-reference control feeds the reference-conditioned teacher a shuffled authoritative answer from another training example, while keeping the student rollout, original prompt, teacher, optimizer, and matched ACE anchors unchanged.
This preserves the reference pathway and context length but removes semantic alignment between the input and the teacher-side reference.
EVOKE Inj. Acc. drops to 18.5, indicating that matched reference content is important for the RoCo likelihood contrast.
Ret. Avg. remains high, and injection does not collapse to the base level because ACE still provides sparse correction for omitted or weakly matched authoritative anchors, while the rollout-distillation path remains anchored to the original prompt.

\textbf{ACE} addresses the complementary omission case.
SDFT+ACE reaches 25.9 EVOKE Inj. Acc., showing that sparse anchored correction helps even without same-rollout likelihood contrast.
ACE only reaches 25.1 EVOKE Inj. Acc. but lowers Ret. Avg. to 53.6, suggesting that anchor correction alone improves injection but lacks the rollout-distillation signal needed to limit drift.
The full \textbf{RoCo-ACE} objective combines rollout-side reweighting with reference-side anchor correction, reaching the best EVOKE Inj. Acc. of 27.6 while keeping Ret. Avg. at 56.5.

The full-reference CE controls show why anchor-level correction matters.
Here, SFT applies uniform CE over the complete authoritative answer, while SDFT+SFT and \textbf{RoCo}+SFT add the same full-reference CE term to the corresponding online-distillation objectives.
SDFT+SFT and \textbf{RoCo}+SFT improve injection less effectively than their ACE counterparts and lower Ret. Avg. to 53.0 and 54.4, respectively.
This shows that the gains do not come merely from adding ordinary supervised CE.
Anchor-level correction is more favorable than full-answer token-level imitation because it restricts supervised loss to extracted factual spans rather than generic reference wording or answer style.

\subsection{Qualitative Example}
Figure~\ref{fig:main_qualitative_case} shows an EVOKE example from the training logs. Additional cases are provided in Appendix~\ref{sec:qualitative_analysis}.
The rollout reaches the right visual and topical neighborhood, mentioning Amsterdam, NDSM, street art, and graffiti, but omits the target museum identity and several concrete facts.
\textbf{RoCo} assigns larger weights to these reference-supported rollout tokens, while \textbf{ACE} adds sparse anchored correction for omitted authoritative anchors such as ``STRAAT Museum'', ``October 9, 2020'', and ``86,000 sq ft''.
This example illustrates the intended split: rollout-side reweighting for content already reached on-policy, and reference-side correction for omitted facts.

\begin{figure}[!t]
\centering
\qualbox{PromptBorder}{PromptBg}{0.93\columnwidth}{Prompt}{Could you explain the details of the museum featured in the image?\par\caseimage[0.50\columnwidth]{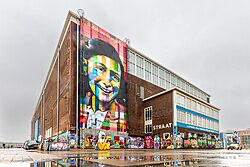}}
\vspace{2pt}

\qualbox{StudentBorder}{StudentBg}{0.93\columnwidth}{Student rollout}{The model identifies a different Amsterdam museum and mixes unsupported history about a warehouse and cooperative organizations. It nevertheless stays close to the visual category by mentioning Amsterdam, NDSM, street art, graffiti, and a cultural venue.}
\vspace{2pt}

\qualbox{ReferenceBorder}{ReferenceBg}{0.93\columnwidth}{Authoritative reference}{The STRAAT Museum is an Amsterdam NDSM street-art and graffiti museum opened on October 9, 2020. Its 86{,}000 sq ft space contains over 180 works from more than 170 artists, includes a mezzanine gallery, and features exterior murals such as Eduardo Kobra's 2016 Anne Frank depiction.}
\vspace{2pt}

\qualbox{RecoBorder}{RecoBg}{0.93\columnwidth}{RoCo signal}{High-weight rollout spans include \textcolor{RecoBorder}{``museum in Amsterdam''}, \textcolor{RecoBorder}{``NDSM''}, \textcolor{RecoBorder}{``street art''}, and \textcolor{RecoBorder}{``graffiti''}. These partial tokens preserve useful visual and topical context.}
\vspace{2pt}

\qualbox{AnchorBorder}{AnchorBg}{0.93\columnwidth}{ACE correction}{Missing anchors include \textcolor{AnchorBorder}{``STRAAT Museum''}, \textcolor{AnchorBorder}{``October 9, 2020''}, \textcolor{AnchorBorder}{``86{,}000 sq ft''}, \textcolor{AnchorBorder}{``over 180 works''}, \textcolor{AnchorBorder}{``more than 170 artists''}, \textcolor{AnchorBorder}{``mezzanine gallery''}, and \textcolor{AnchorBorder}{``Eduardo Kobra''}.}
\caption{
Representative EVOKE example illustrating the complementarity between \textbf{RoCo} reweighting and \textbf{ACE} anchored correction.
\textbf{RoCo} emphasizes reference-supported rollout tokens already reached on-policy, while \textbf{ACE} supplies sparse correction for authoritative anchors omitted from the rollout.
}
\vspace{-8pt}
\label{fig:main_qualitative_case}
\end{figure}



\section{Conclusion}
We presented \textbf{RoCo-ACE}, a rollout-conditioned contrastive online distillation objective for knowledge injection.
\textbf{RoCo} uses same-rollout reference-free/reference-conditioned likelihood contrast to reweight reference-supported rollout tokens, while \textbf{ACE} adds sparse anchored correction for authoritative anchors omitted from or weakly matched by the rollout.
Together, the two objectives inject sparse factual content without fitting the full reference answer as a token-level imitation target.
Experiments across three knowledge-injection settings and multiple retention benchmarks show improved injected-knowledge accuracy while keeping evaluated retention close to the base model.


\section*{Limitations}
\textbf{RoCo-ACE} is designed for knowledge-injection settings where each update is accompanied by an authoritative natural-language reference.
Its supervision is organized around student rollout tokens and extracted factual anchors, making it most directly suited to entity, event, domain, and visual-world updates whose target facts can be expressed as compact text spans.
In label-only settings, very short answers, or cases where the relevant evidence is mostly implicit in the image, the available anchor signal becomes weaker.

Although images are included in both student and teacher contexts, the current objective does not use region-level, OCR-level, or object-level evidence annotations to explicitly ground each corrected anchor.
A natural extension is to tie anchored factual correction more directly to visual evidence through image-ablation contrast, OCR/entity grounding, or region-level supervision.
Finally, retention is evaluated empirically on held-out multimodal and safety benchmarks; the method is not intended to provide formal guarantees over all non-updated capabilities or deployment distributions.






\clearpage
\appendix

\section{Anchor Extraction and Supervision-Weight Diagnostics}
\label{sec:span_coverage}

\begin{table}[!h]
\centering

\scriptsize
\setlength{\tabcolsep}{2pt}
\renewcommand{\arraystretch}{1.08}
\resizebox{\columnwidth}{!}{
\begin{tabular}{@{}p{0.27\columnwidth}cp{0.22\columnwidth}p{0.38\columnwidth}@{}}
\toprule
\textbf{Token category} & \textbf{Share} & \textbf{Component} & \textbf{Diagnostic rule} \\
\midrule
Reference-supported rollout tokens & 30.8\% & RoCo enhanced KD & Valid rollout tokens satisfying $\Delta_{i,t}>\tau$. These positions receive additional RoCo contrast weight. \\
Generic or weakly supported rollout tokens & 57.3\% & RoCo KD floor & Valid rollout tokens without RoCo contrast activation; these positions are mainly retained by the online-distillation floor. \\
Missing target-anchor tokens & 11.9\% & ACE correction & Extracted authoritative-anchor tokens absent from the rollout; these positions receive ACE missing-span correction. \\
\bottomrule
\end{tabular}
}
\caption{Token-allocation diagnostics on EVOKE training rollouts. The diagnostic pool consists of valid student-rollout tokens plus missing reference-anchor tokens. The three categories are mutually exclusive and sum to one, showing how RoCo-ACE allocates RoCo and ACE supervision across reference-supported rollout content, generic rollout content, and missing target anchors. Percentages are computed from logged RoCo-ACE training intervals under the paper's default anchor weighting.}
\label{tab:token_weight_diagnostics}
\end{table}


This section describes the lightweight span extraction used by \textbf{ACE} and RoCo span smoothing, and reports a supervision-weight diagnostic that connects these extracted units to the claims in the main paper.
The diagnostic is not intended to estimate final gradient norms or normalized loss mass.
Instead, it shows how training units are assigned to the RoCo contrast term, the RoCo floor term, and reference-side ACE correction.

\paragraph{Span extraction.}
Span construction is a tokenizer-level procedure and does not rely on external NER or parsing tools.
We decode valid response tokens, remove special tokens, punctuation-only tokens, and common function words, and group contiguous content tokens containing CJK characters, Latin letters, or digits into spans.
A span is kept if it contains at least two visible characters or any digit, and we cap each span by a maximum token length.
This simple extractor is designed to capture compact factual expressions such as entity names, dates, numbers, titles, locations, and short noun phrases.

\paragraph{Anchor matching for ACE.}
For ACE, the same extractor is applied to the tokenized authoritative answer.
Each reference span is normalized by lowercasing and retaining only alphanumeric and CJK characters; the student rollout is normalized in the same way.
For each extracted reference span $s$, we compute an anchor match score $m_i(s,y_i)\in[0,1]$ against the rollout.
In our default implementation, $m_i(s,y_i)=1$ when the normalized span string appears in the normalized rollout and $0$ otherwise; the same notation also supports partial-match variants.
Thus, matched anchors receive the base anchor weight, while omitted or weakly matched anchors receive larger weights.
The ACE token set $\mathcal{A}_i$ is the union of reference-token positions covered by the extracted anchors.
Tokens outside $\mathcal{A}_i$ serve only as teacher-forced context and do not receive supervised CE loss, making ACE narrower than full-reference SFT.

\paragraph{Span smoothing for RoCo.}
The same extractor is also used for RoCo span-aware smoothing on the rollout side.
After preliminary RoCo weights are computed from the same-rollout likelihood contrast, we aggregate token weights within each extracted rollout span and broadcast the span score back to all tokens in that span.
For notation, the resulting smoothed weights are still denoted by $w_{i,t}$.
This reduces subword-level noise for names, dates, numbers, multilingual expressions, and other compact factual units without requiring external entity taggers.
The diagnostic in Table~\ref{tab:token_weight_diagnostics} classifies rollout tokens by their pre-smoothing contrast activation, so it reflects the source of the RoCo weight signal before span-level broadcasting.

\paragraph{Supervision-weight diagnostic.}
Table~\ref{tab:token_weight_diagnostics} reports a token-level diagnostic on EVOKE training rollouts.
The diagnostic pool contains valid rollout tokens and omitted authoritative-anchor tokens.
These units correspond to the three supervision cases used by \textbf{RoCo-ACE}: rollout tokens receiving the RoCo contrast term, rollout tokens receiving mainly the RoCo floor term, and reference-side anchors receiving ACE omission correction.

For rollout tokens, Table~\ref{tab:token_weight_diagnostics} uses the RoCo likelihood contrast as the diagnostic rule.
Tokens with $\Delta_{i,t}>\tau$ are counted as reference-supported rollout tokens; they account for 30.8\% of the diagnostic pool and receive the RoCo floor plus the additional contrast term.
Tokens with $\Delta_{i,t}\le\tau$ are counted as generic or non-activated rollout tokens; they account for 57.3\% and mainly receive the online-distillation floor.
For reference-side units, omitted authoritative-anchor tokens account for 11.9\% and receive larger ACE weights through the omission term $1-\rho_i(s,y_i)$.

This distribution supports the intended supervision split.
Uniform reference-conditioned rollout distillation would place its rollout-side loss over all valid rollout tokens, most of which are not contrast-activated in this diagnostic.
RoCo instead reallocates additional distillation weight to reference-supported rollout tokens, while ACE supplies sparse correction for omitted authoritative anchors that are not available as generated rollout tokens in the current step.
Because Table~\ref{tab:token_weight_diagnostics} reports token-count categories rather than normalized loss mass, it should be interpreted as a weight-assignment diagnostic rather than an exact gradient-budget decomposition.

\section{Full Baseline Comparison Results}
\label{app:full_baseline_results}


\begin{figure}
    \includegraphics[width=\linewidth]{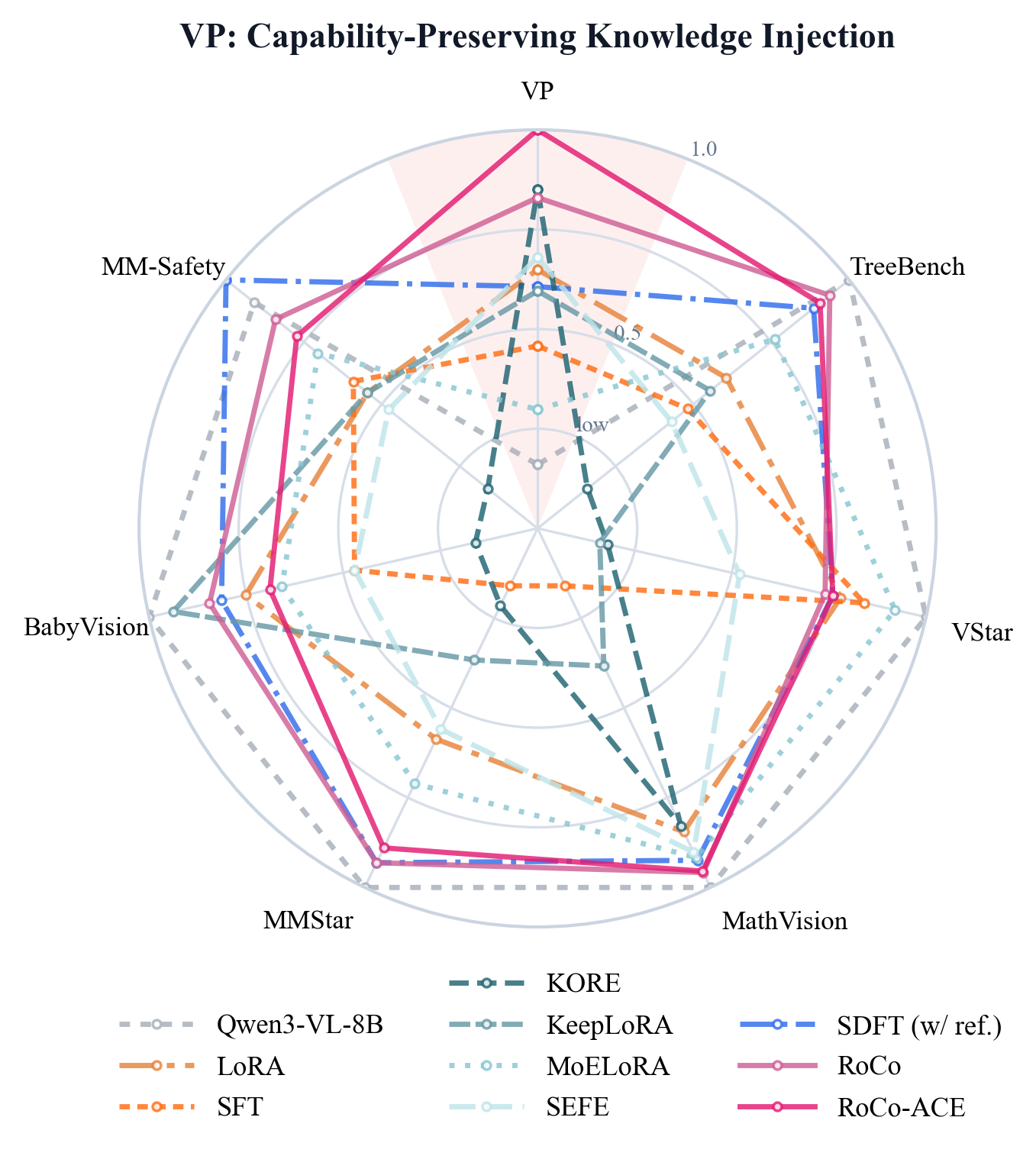}
    \captionof{figure}{Radar comparison on VP, showing target knowledge injection and general capability retention across six multimodal benchmarks.}
    \label{fig:main_baseline_results_vp}
\end{figure}
    
\begin{figure}
    \includegraphics[width=\linewidth]{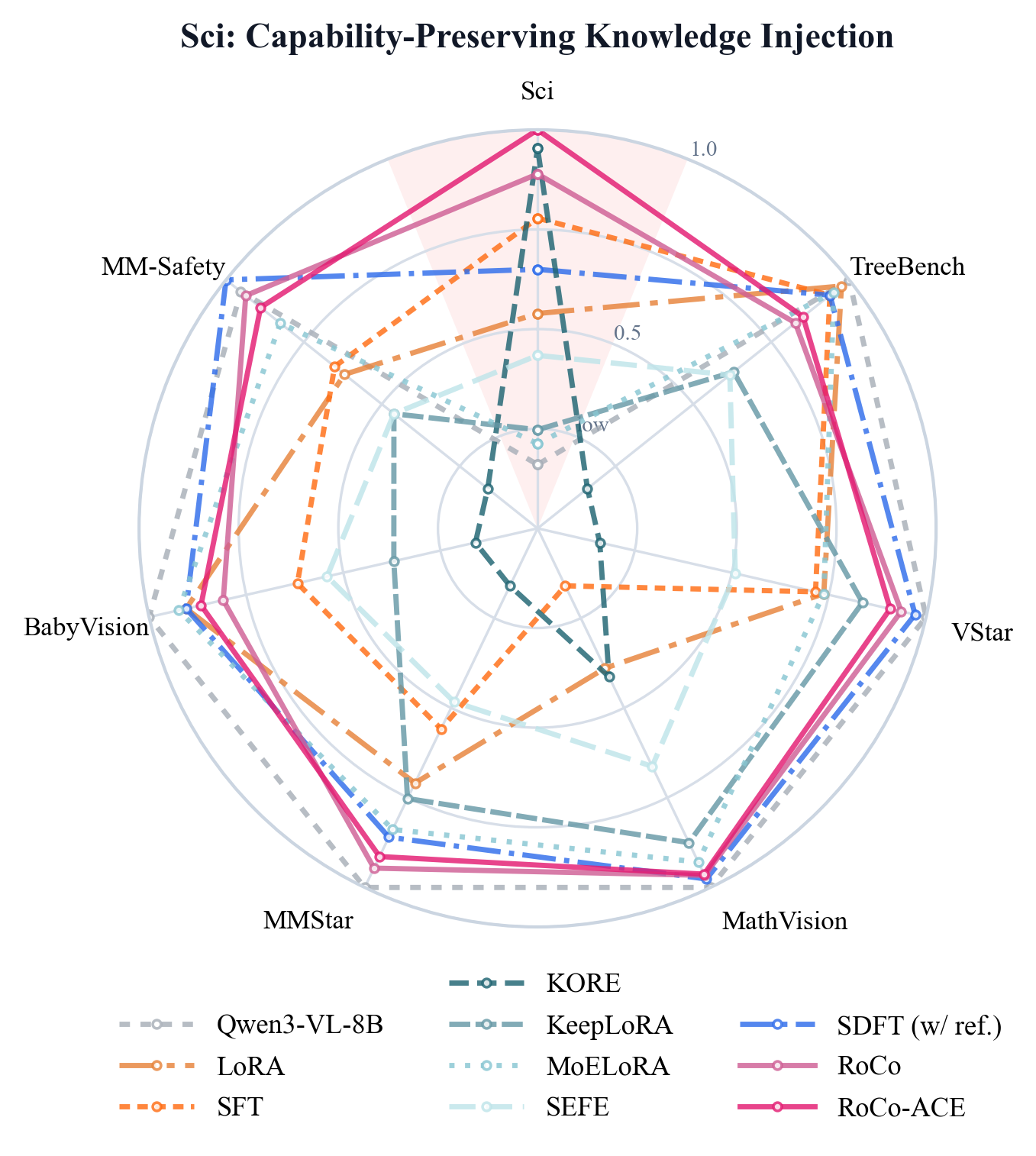}
    \captionof{figure}{Radar comparison on Sci, showing target knowledge injection and general capability retention across six multimodal benchmarks.}
    \label{fig:main_baseline_results_sci}
\end{figure}

Table~\ref{tab:main_baseline_results} reports the full baseline comparison on EVOKE using Qwen3-VL-8B as the base model. In addition to the injected-knowledge accuracy, we include all individual retention benchmarks to show how each method affects general multimodal capabilities after knowledge injection. This table complements the main-text discussion by making the injection--retention trade-off explicit across direct fine-tuning, constrained-update baselines, online distillation, and our RoCo-ACE variants.

\begin{table*}[!t]
\centering

\small
\setlength{\tabcolsep}{3.8pt}
\resizebox{\textwidth}{!}{
\begin{tabular}{lcccccccc}
\toprule
\multirow{2}{*}{\textbf{Method}} &
\multicolumn{1}{c}{\textbf{Knowledge Injection}} &
\multicolumn{7}{c}{\textbf{Retention Benchmark}} \\
\cmidrule(lr){2-2}\cmidrule(lr){3-9}
& \textbf{EVOKE}
& \textbf{TreeBench} & \textbf{VStar} & \textbf{MathVision} & \textbf{MMStar} & \textbf{BabyVision} & \textbf{MM-Safety} & \textbf{Ret. Avg.} \\
\midrule
Qwen3-VL-8B & 16.4 & \textbf{43.3} & 83.2 & \textbf{49.3} & \textbf{70.6} & \textbf{14.2} & \textbf{82.6} & \textbf{57.2} \\
\midrule
LoRA & 18.6 & 26.2 & 70.9 & 35.6 & 59.7 & 13.5 & 65.3 & 45.2 \\
SFT & 20.1 & 15.8 & 68.1 & 29.8 & 35.5 & 10.7 & 30.9 & 31.8 \\
\midrule
SDFT (w/ ref.) & 20.9 & 39.9 & 82.3 & 45.8 & 68.9 & 12.6 & 80.8 & 55.1 \\
\midrule
KORE & \textbf{\textit{25.2}} & 30.6 & 71.9 & 37.8 & 60.6 & 11.5 & 69.8 & 47.0 \\
KeepLoRA & 17.3 & 41.2 & \textbf{84.0} & \textbf{\textit{49.2}} & 70.0 & \textbf{\textit{13.9}} & 81.1 & \textbf{\textit{56.6}} \\
MoELoRA & 17.4 & 32.5 & 73.1 & 38.6 & 63.5 & 13.1 & 71.5 & 48.7 \\
SEFE & 19.8 & 31.0 & 70.5 & 32.4 & 56.7 & 11.2 & 66.7 & 44.8 \\
\midrule
\textbf{RoCo} & 23.7 & \textbf{\textit{42.9}} & 82.9 & 48.1 & 69.9 & 13.8 & \textbf{\textit{81.7}} & \textbf{\textit{56.6}} \\
\textbf{RoCo-ACE} & \textbf{27.6} & 41.6 & \textbf{\textit{83.9}} & 48.2 & \textbf{\textit{70.1}} & 13.5 & 81.5 & 56.5 \\
\bottomrule
\end{tabular}
}
\caption{Main baseline comparison on Qwen3-VL-8B. The knowledge-injection column reports LLM-judge accuracy on EVOKE. Retention benchmark columns report accuracy on held-out general multimodal and safety benchmarks, and Ret. Avg. averages TreeBench, VStar, MathVision, MMStar, BabyVision, and MM-Safety. Higher values are better; bold and bold-italic mark the best and second-best scores in each column.}
\label{tab:main_baseline_results}
\end{table*}


Table~\ref{tab:additional_injection_results} further evaluates the same set of methods on the VP and Sci knowledge-injection benchmarks from MLLM-CL. VP focuses on ability-oriented injection, while Sci focuses on domain-oriented knowledge injection. We keep the retention benchmark suite unchanged across EVOKE, VP, and Sci, which allows a consistent comparison of whether each method can acquire new knowledge while preserving general capabilities. Across these additional settings, RoCo-ACE maintains a favorable balance between injected-knowledge accuracy and retention performance.

\begin{table*}[!t]
\centering

\small
\setlength{\tabcolsep}{2.4pt}
\resizebox{\textwidth}{!}{
\begin{tabular}{llcccccccc}
\toprule
\multirow{2}{*}{\textbf{Dataset}} &
\multirow{2}{*}{\textbf{Method}} &
\multicolumn{1}{c}{\textbf{Injection}} &
\multicolumn{7}{c}{\textbf{Retention Benchmarks}} \\
\cmidrule(lr){3-3}\cmidrule(lr){4-10}
& & \textbf{Acc.}
& \textbf{TreeBench} & \textbf{VStar} & \textbf{MathVision} & \textbf{MMStar} & \textbf{BabyVision} & \textbf{MM-Safety} & \textbf{Ret. Avg.} \\
\midrule
\multirow{10}{*}{VP}
& Qwen3-VL-8B & 69.9 & \textbf{43.3} & \textbf{83.2} & \textbf{49.3} & \textbf{70.6} & \textbf{14.2} & \textbf{\textit{82.6}} & \textbf{57.2} \\
& LoRA & 74.5 & 39.5 & 82.1 & 45.2 & 67.6 & 13.4 & 81.0 & 54.8 \\
& SFT & 72.7 & 38.3 & 82.4 & 27.1 & 64.5 & 12.5 & 81.2 & 51.0 \\
& SDFT (w/ ref.) & 74.1 & 42.2 & 82.0 & 47.3 & \textbf{\textit{70.1}} & 13.6 & \textbf{83.0} & 56.4 \\
& KORE & \textbf{\textit{76.4}} & 35.2 & 79.1 & 44.8 & 64.9 & 11.5 & 79.3 & 52.5 \\
& KeepLoRA & 74.0 & 39.0 & 79.0 & 33.0 & 66.0 & \textbf{\textit{14.0}} & 81.0 & 52.0 \\
& MoELoRA & 71.2 & 41.0 & \textbf{\textit{82.8}} & 47.1 & 68.5 & 13.1 & 81.7 & 55.7 \\
& SEFE & 74.8 & 37.8 & 80.8 & 46.7 & 67.4 & 12.5 & 80.7 & 54.3 \\
& \textbf{RoCo} & 76.2 & \textbf{\textit{42.7}} & 81.9 & \textbf{\textit{48.2}} & \textbf{\textit{70.1}} & 13.7 & 82.3 & \textbf{\textit{56.5}} \\
& \textbf{RoCo-ACE} & \textbf{77.8} & 42.4 & 82.0 & 48.1 & 69.8 & 13.2 & 82.0 & 56.2 \\
\midrule
\multirow{10}{*}{Sci}
& Qwen3-VL-8B & 73.1 & \textbf{43.3} & \textbf{83.2} & \textbf{49.3} & \textbf{70.6} & \textbf{14.2} & \textbf{\textit{82.6}} & \textbf{57.2} \\
& LoRA & 79.6 & \textbf{\textit{43.1}} & 79.5 & 31.2 & 67.9 & 13.7 & 80.5 & 52.6 \\
& SFT & 83.7 & 42.8 & 79.2 & 24.4 & 66.5 & 12.2 & 80.7 & 51.0 \\
& SDFT (w/ ref.) & 81.5 & 42.8 & \textbf{\textit{82.8}} & \textbf{\textit{48.6}} & 69.3 & 13.7 & \textbf{82.9} & \textbf{\textit{56.7}} \\
& KORE & \textbf{\textit{86.7}} & 36.5 & 71.4 & 31.9 & 62.8 & 9.8 & 77.6 & 48.3 \\
& KeepLoRA & 74.6 & 40.3 & 80.9 & 45.6 & 68.3 & 10.9 & 79.5 & 54.2 \\
& MoELoRA & 74.0 & 42.9 & 79.5 & 47.2 & 69.1 & \textbf{\textit{13.8}} & 81.8 & 55.7 \\
& SEFE & 77.8 & 40.2 & 76.3 & 39.3 & 65.8 & 11.8 & 79.5 & 52.1 \\
& \textbf{RoCo} & 85.6 & 41.9 & 82.3 & 48.3 & \textbf{\textit{70.1}} & 13.2 & 82.5 & 56.4 \\
& \textbf{RoCo-ACE} & \textbf{87.5} & 42.1 & 81.9 & 48.2 & 69.8 & 13.5 & 82.2 & 56.3 \\
\bottomrule
\end{tabular}
}
\caption{Baseline comparison on the MLLM-CL VP and Sci knowledge-injection benchmarks \citep{zhao2025mllmcl} with full retention metrics. Acc. reports injected-knowledge accuracy for the corresponding dataset. Retention columns report accuracy after training on that dataset, and Ret. Avg. averages TreeBench, VStar, MathVision, MMStar, BabyVision, and MM-Safety. Higher values are better; bold and bold-italic mark the best and second-best scores in each column.}
\label{tab:additional_injection_results}
\end{table*}


Figures~\ref{fig:main_baseline_results_vp} and~\ref{fig:main_baseline_results_sci} provide the corresponding radar visualizations for VP and Sci. We move these plots to the appendix because the main text summarizes the three datasets in Table~\ref{tab:main_three_dataset_summary} using injected-knowledge accuracy and Ret. Avg.


\section{Qualitative Analysis}
\label{sec:qualitative_analysis}

\begin{table*}[!t]
\centering

\setlength{\tabcolsep}{3pt}
\begin{tabular}{p{0.47\textwidth}p{0.47\textwidth}}
\toprule
\textbf{Case 1: generic under-injection in a business entity} &
\textbf{Case 2: sports news numeric/event correction} \\
\midrule
\casebox{PromptBorder}{PromptBg}{Prompt}{What details can you give about the business in the image?\par\caseimage[0.32\linewidth]{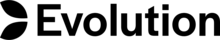}}
\vspace{2pt}

\casebox{ReferenceBorder}{ReferenceBg}{Authoritative reference}{Evolution AB is a Swedish gaming technology company based in Stockholm, specializing in B2B live casino software for online operators. It was founded in 2006 by Jens von Bahr and Fredrik \"{O}sterberg, became publicly listed on Nasdaq in 2015, introduced games such as Monopoly Live and Lightning Roulette, acquired Ezugi and NetEnt, and attracted investment from Kenneth Dart in 2024.}
\vspace{2pt}

\casebox{StudentBorder}{StudentBg}{Student rollout}{The rollout gives a generic description of startups, growth, innovation, vision clarity, branding, customer loyalty, and market adaptation. It never identifies Evolution AB or its concrete business facts.}
\vspace{2pt}

\casebox{RecoBorder}{RecoBg}{\textbf{RoCo} signal}{The strongest rollout signals are generic business phrases such as ``Startups abound with growth and innovation'' and ``successful startups have a crystal-clear mission''. This case exposes a limitation of rollout-only reweighting: without the target entity in the rollout, the contrast cannot directly teach Evolution AB.}
\vspace{2pt}

\casebox{AnchorBorder}{AnchorBg}{ACE correction}{Missing spans include ``Evolution AB'', ``Swedish gaming technology company'', ``Stockholm'', ``B2B live casino software'', ``Jens von Bahr'', ``Fredrik \"{O}sterberg'', ``Nasdaq in 2015'', ``Monopoly Live'', ``Lightning Roulette'', ``Ezugi'', ``NetEnt'', and ``Kenneth Dart in 2024''.}
&
\casebox{PromptBorder}{PromptBg}{Prompt}{What information can you offer about the sport news presented in the image?\par\caseimage[0.32\linewidth]{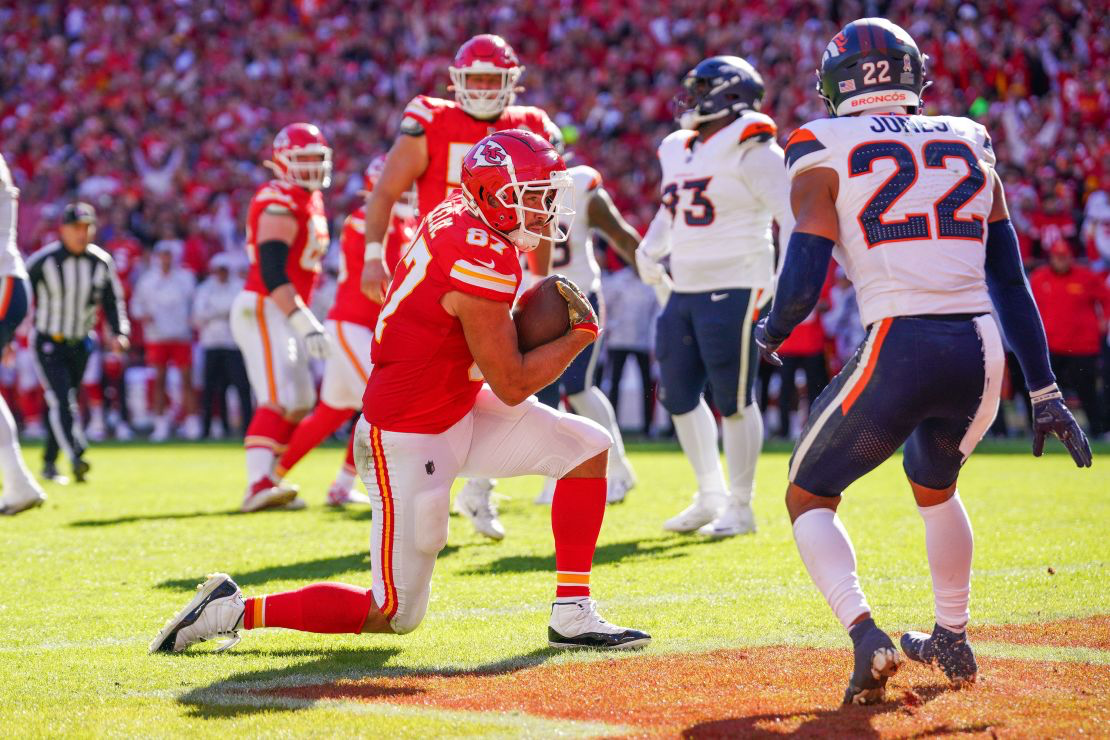}}
\vspace{2pt}

\casebox{ReferenceBorder}{ReferenceBg}{Authoritative reference}{The Kansas City Chiefs defeated the Denver Broncos 16--14 at Arrowhead Stadium on November 10, 2024, blocking a last-second field goal to preserve their unbeaten streak. Patrick Mahomes led a late-game drive, Leo Chenal made the decisive block, and Travis Kelce set a franchise touchdown-reception record.}
\vspace{2pt}

\casebox{StudentBorder}{StudentBg}{Student rollout}{The rollout reaches the right sports-news neighborhood: it mentions Travis Kelce, the Kansas City Chiefs, the Denver Broncos, Arrowhead Stadium, and a narrow victory. However, it shifts the event to a season opener or preseason game, gives an imprecise three-point outcome, and introduces unsupported players and game details.}
\vspace{2pt}

\casebox{RecoBorder}{RecoBg}{\textbf{RoCo} focus}{High-weight rollout spans include ``Travis Kelce and the Kansas City Chiefs'', ``victory against the Denver Broncos'', ``Arrowhead Stadium'', and ``brief summary of the news''. These tokens preserve useful event context rather than treating the entire generated answer uniformly.}
\vspace{2pt}

\casebox{AnchorBorder}{AnchorBg}{ACE correction}{Missing or contradicted spans include ``16--14'', ``last-second field goal attempt'', ``November 10, 2024'', ``Patrick Mahomes'', ``Leo Chenal'', ``franchise record'', ``Taylor Swift'', ``15th consecutive win'', and ``Buffalo Bills''.}
\\
\midrule
\multicolumn{2}{p{0.94\textwidth}}{\textbf{Case 3: India news collaborator substitution}} \\
\midrule
\multicolumn{2}{p{0.94\textwidth}}{
\casebox{PromptBorder}{PromptBg}{Prompt}{Can you please share some details about the india news shown in the image?\par\caseimage[0.32\linewidth]{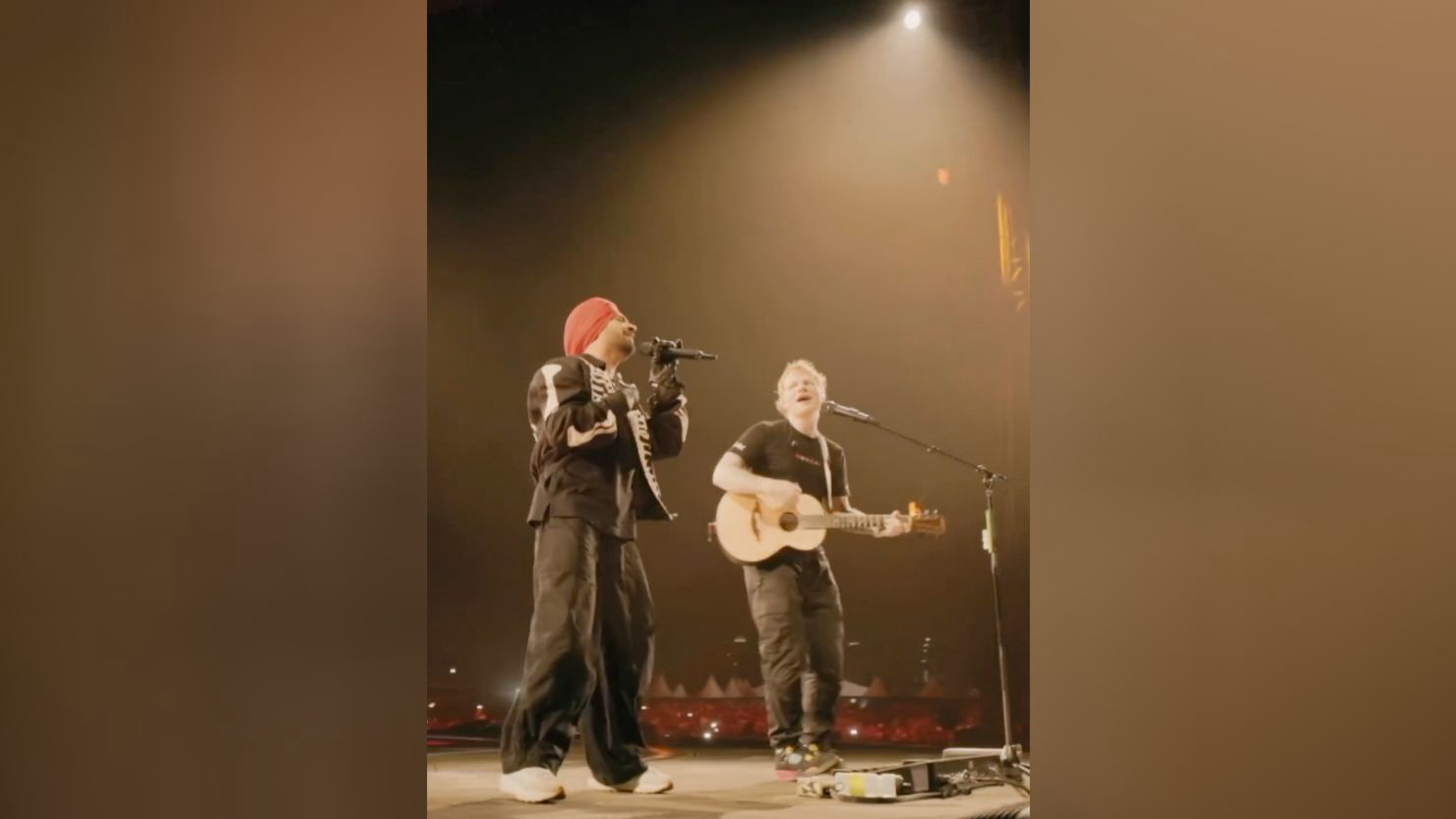}}
\vspace{2pt}

\casebox{ReferenceBorder}{ReferenceBg}{Authoritative reference}{Ed Sheeran performed a duet with Punjabi star Diljit Dosanjh in Mumbai before a 50{,}000-strong audience, singing a Punjabi rendition of Dosanjh's hit ``Lover''. The surprise collaboration at Mahalaxmi Racecourse drew social-media attention during Sheeran's ``Mathematics'' tour.}
\vspace{2pt}

\casebox{StudentBorder}{StudentBg}{Student rollout}{The rollout remains in the broad music-news setting and mentions Ed Sheeran, Punjabi music, collaboration, crowds, and concerts, but it substitutes the collaborator with Sidhu Moose Wala and introduces unrelated claims about surveillance and cancelled tours.}
\vspace{2pt}

\casebox{RecoBorder}{RecoBg}{\textbf{RoCo} focus}{High-weight spans include ``summary of the news'', ``British singer Ed Sheeran'', ``Punjabi'', ``joint performances'', and ``large crowds'', so online distillation is concentrated on the useful music-news neighborhood.}
\vspace{2pt}

\casebox{AnchorBorder}{AnchorBg}{ACE correction}{Missing or contradicted spans include ``Diljit Dosanjh'', ``Mumbai'', ``50{,}000-strong audience'', ``Punjabi rendition of Dosanjh's hit Lover'', ``Mahalaxmi Racecourse'', ``Mathematics tour'', and ``performing in Punjabi for the first time''.}
}
\\
\bottomrule
\end{tabular}
\caption{Qualitative EVOKE examples from joint \textbf{RoCo-ACE} training. We retain three non-duplicate cases that cover entity under-injection, sports-event correction, and collaborator substitution. Across these cases, \textbf{RoCo} emphasizes useful content already present in the rollout, while ACE supplies authoritative facts that the rollout omits or contradicts.}
\label{tab:qualitative_cases}
\end{table*}


\begin{table*}[!t]
\centering
\refstepcounter{subsection}\label{sec:baseline_family_examples}
\setlength{\tabcolsep}{3pt}
\begin{tabular}{p{0.47\textwidth}p{0.47\textwidth}}
\toprule
\multicolumn{2}{p{0.94\textwidth}}{\textbf{\thesubsection\quad Baseline-Family Behavior Examples.}
Representative EVOKE cases illustrating how baseline families behave on sparse authoritative references: direct fitting captures a few salient facts, constrained updating covers a moderate subset, online distillation preserves natural response style but under-injects, and \textbf{RoCo-ACE} recovers more anchors without copying the full reference.} \\
\midrule
\textbf{Case A: album entity with release and credit facts} &
\textbf{Case B: film entity with series, cast, and release facts} \\
\midrule
\casebox{PromptBorder}{PromptBg}{Question}{Could you introduce the album in the image?\par\caseimage[0.30\linewidth]{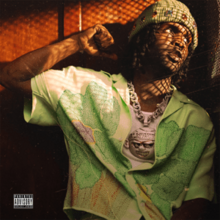}}
\vspace{2pt}

\casebox{ReferenceBorder}{ReferenceBg}{Gold facts}{\textbf{Almighty So 2} is the \textbf{fifth studio album} by \textbf{Chief Keef}, released on \textbf{May 10, 2024} by \textbf{43B}. It is a \textbf{sequel to Almighty So}, includes guests such as \textbf{G Herbo, Tierra Whack, Sexyy Red, and Quavo}, has production by Keef and several collaborators, and peaked at \textbf{number 30 on the Billboard 200}.}
\vspace{2pt}

\casebox{StudentBorder}{StudentBg}{SFT / direct fitting}{Captures \textbf{Almighty So 2}, \textbf{Chief Keef}, and a 2024 album identity, but gives little detail about the label, guest artists, production, or chart performance.}
\vspace{2pt}

\casebox{RecoBorder}{RecoBg}{KORE / constrained update}{Adds the \textbf{fifth studio album}, \textbf{May 10, 2024}, and \textbf{43B} facts, but still only briefly covers guests and omits most production and chart details.}
\vspace{2pt}

\casebox{ReferenceBorder}{ReferenceBg}{SDFT / online distillation}{Keeps natural album-cover wording and visual grounding, but avoids firm claims about the exact \textbf{release label, guest list, producers, and Billboard ranking}.}
\vspace{2pt}

\casebox{AnchorBorder}{AnchorBg}{\textbf{RoCo-ACE} / rollout-conditioned anchors}{Recovers \textbf{Almighty So 2}, \textbf{fifth studio album}, \textbf{Chief Keef}, \textbf{May 10, 2024}, \textbf{43B}, several guest artists, and \textbf{number 30 on the Billboard 200}, while not copying every production detail.}
&
\casebox{PromptBorder}{PromptBg}{Question}{Can you share information about the film presented in the image?\par\caseimage[0.30\linewidth]{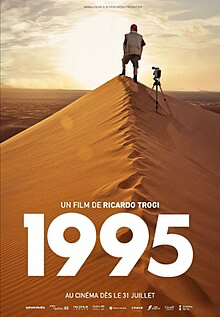}}
\vspace{2pt}

\casebox{ReferenceBorder}{ReferenceBg}{Gold facts}{\textbf{1995} is a \textbf{2024 Canadian comedy film} written and directed by \textbf{Ricardo Trogi}. It is the \textbf{fourth film} in his semi-autobiographical series after \textbf{1981, 1987, and 1991}, stars \textbf{Jean-Carl Boucher as Trogi}, and was released on \textbf{July 31, 2024} by \textbf{Immina Films}.}
\vspace{2pt}

\casebox{StudentBorder}{StudentBg}{SFT / direct fitting}{Captures \textbf{1995}, \textbf{Canadian comedy film}, and \textbf{Ricardo Trogi}, but gives little information about the cast, release date, distributor, or story setup.}
\vspace{2pt}

\casebox{RecoBorder}{RecoBg}{KORE / constrained update}{Adds \textbf{2024}, the semi-autobiographical series relation, and \textbf{Jean-Carl Boucher}, but simplifies the premise and omits distributor details.}
\vspace{2pt}

\casebox{ReferenceBorder}{ReferenceBg}{SDFT / online distillation}{Preserves natural film-poster description, but remains cautious about the exact \textbf{series order, cast, release date, and distributor}.}
\vspace{2pt}

\casebox{AnchorBorder}{AnchorBg}{\textbf{RoCo-ACE} / rollout-conditioned anchors}{Covers \textbf{1995}, \textbf{2024 Canadian comedy}, \textbf{Ricardo Trogi}, the \textbf{fourth-entry} relation, \textbf{Jean-Carl Boucher as Trogi}, and \textbf{July 31, 2024}, while leaving out some supporting cast details.}
\\
\bottomrule
\end{tabular}
\end{table*}

\begin{table*}[!t]
\centering

\setlength{\tabcolsep}{3pt}
\begin{tabular}{p{0.94\textwidth}}
\toprule
\textbf{Case C: person entity with sports, medical, and family facts} \\
\midrule
\casebox{PromptBorder}{PromptBg}{Question}{Can you describe the human depicted in the image?\par\caseimage[0.20\textwidth]{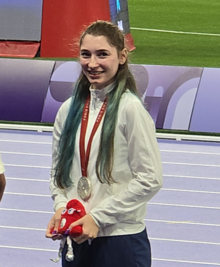}}
\vspace{2pt}

\casebox{ReferenceBorder}{ReferenceBg}{Gold facts}{\textbf{Lida-Maria Manthopoulou}, born on \textbf{June 14, 2005}, is a \textbf{Greek para-athlete} who competed in the \textbf{2024 Summer Paralympics} and won \textbf{silver in the 100 metres T38}. She represents \textbf{Elpides Thessaloniki}, was diagnosed with \textbf{multiple sclerosis} before the Paralympics, and is connected to actor \textbf{Aias Manthopoulos} and other public family members.}
\vspace{2pt}

\casebox{StudentBorder}{StudentBg}{SFT / direct fitting}{Captures \textbf{Lida-Maria Manthopoulou}, Greek para-athlete identity, and a 2024 Paralympics association, but omits most event, medal, club, health, and family facts.}
\vspace{2pt}

\casebox{RecoBorder}{RecoBg}{KORE / constrained update}{Adds the \textbf{silver medal} and \textbf{100 metres T38} event, but only briefly mentions background and omits most medical and family details.}
\vspace{2pt}

\casebox{ReferenceBorder}{ReferenceBg}{SDFT / online distillation}{Keeps natural person-description style, but avoids firm details about the \textbf{Paralympic event, medal, diagnosis, club, and family lineage}.}
\vspace{2pt}

\casebox{AnchorBorder}{AnchorBg}{\textbf{RoCo-ACE} / rollout-conditioned anchors}{Recovers the person identity, \textbf{Greek para-athlete}, \textbf{June 14, 2005}, \textbf{2024 Summer Paralympics}, \textbf{silver in the 100 metres T38}, \textbf{Elpides Thessaloniki}, and the \textbf{Aias Manthopoulos} family link, while leaving out some extended family details.}
\\
\bottomrule
\end{tabular}
\caption{Baseline-family behavior examples (continued).}
\end{table*}


Table~\ref{tab:qualitative_cases} presents three non-duplicate EVOKE cases selected from the joint \textbf{RoCo-ACE} training logs. The color scheme follows the method overview: orange boxes summarize the student rollout, green boxes show the strongest \textbf{RoCo} rollout signals, and red boxes show facts corrected by Anchored Cross-Entropy. Together, these examples show how \textbf{RoCo} emphasizes useful on-policy content while ACE supplies missing or contradicted reference facts.

The baseline-family examples in Appendix~\ref{sec:baseline_family_examples} further illustrate how different method families behave on sparse authoritative references. Direct fitting usually captures only a few salient facts, constrained-update methods cover a moderate subset, online distillation preserves natural response style but under-injects, and \textbf{RoCo-ACE} recovers more anchors without copying the full reference.


\section{Supplementary Results on Qwen3-VL-30B}
\label{sec:qwen3vl30b_results}

\begin{table*}[!t]
\centering

\small
\setlength{\tabcolsep}{2.2pt}
\resizebox{\textwidth}{!}{
\begin{tabular}{llcccccccc}
\toprule
\multirow{2}{*}{\textbf{Group}} &
\multirow{2}{*}{\textbf{Method}} &
\multicolumn{1}{c}{\textbf{Knowledge Injection}} &
\multicolumn{7}{c}{\textbf{Retention Benchmark}} \\
\cmidrule(lr){3-3}\cmidrule(lr){4-10}
& & \textbf{EVOKE}
& \textbf{TreeBench} & \textbf{VStar} & \textbf{MathVision} & \textbf{MMStar} & \textbf{BabyVision} & \textbf{MM-Safety} & \textbf{Ret. Avg.} \\
\midrule
Base model & Qwen3-VL-30B & 21.6 & \textbf{\textit{41.3}} & 83.2 & 54.5 & 72.4 & 16.4 & 82.4 & \textbf{\textit{58.4}} \\
\midrule
Direct fitting & LoRA & 23.1 & 38.1 & 79.5 & 50.6 & 67.4 & 12.7 & 76.4 & 54.1 \\
Direct fitting & SFT & 25.1 & 36.5 & 77.4 & 49.6 & 62.7 & 11.7 & 71.7 & 51.6 \\
Online distillation & SDFT (w/ ref.) & 26.4 & 39.4 & 83.1 & 55.0 & 72.1 & 15.3 & \textbf{83.0} & 58.0 \\
\midrule
Constrained updates & KORE & \textbf{\textit{29.7}} & 39.6 & 79.5 & 49.3 & 65.1 & 13.8 & 70.5 & 53.0 \\
Constrained updates & KeepLoRA & 20.8 & 41.2 & 83.0 & 54.1 & 72.0 & 16.4 & 81.9 & 58.1 \\
Constrained updates & MoELoRA & 22.9 & 39.4 & 80.5 & 51.2 & 68.9 & 13.5 & 78.5 & 55.3 \\
Constrained updates & SEFE & 26.1 & 38.6 & 77.5 & 50.1 & 64.2 & 12.8 & 68.4 & 51.9 \\
\midrule
Ours & \textbf{RoCo} & 28.9 & 40.8 & \textbf{\textit{85.4}} & \textbf{56.8} & \textbf{72.9} & \textbf{\textit{16.7}} & \textbf{\textit{82.9}} & \textbf{59.2} \\
Ours & \textbf{RoCo-ACE} & \textbf{31.2} & \textbf{42.0} & \textbf{85.6} & \textbf{\textit{55.4}} & \textbf{\textit{72.8}} & \textbf{16.8} & 82.7 & \textbf{59.2} \\
\bottomrule
\end{tabular}
}
\caption{Supplementary Qwen3-VL-30B comparison. EVOKE is evaluated with the same LLM-judge protocol as the main table. Retention columns report held-out general multimodal and safety accuracy, and Ret. Avg. averages TreeBench, VStar, MathVision, MMStar, BabyVision, and MM-Safety. Higher values are better; bold and bold-italic mark the best and second-best scores in each column.}
\label{tab:qwen3vl30b_results}
\end{table*}


\begin{table*}[!t]
\centering

\small
\setlength{\tabcolsep}{4.0pt}
\resizebox{\textwidth}{!}{
\begin{tabular}{lcccccccc}
\toprule
\multirow{2}{*}{\textbf{Variant}} &
\multirow{2}{*}{\textbf{EVOKE}} &
\multicolumn{7}{c}{\textbf{Retention Benchmarks}} \\
\cmidrule(lr){3-9}
& & \textbf{TreeBench} & \textbf{VStar} & \textbf{MathVision} & \textbf{MMStar} & \textbf{BabyVision} & \textbf{MM-Safety} & \textbf{Ret. Avg.} \\
\midrule
Base model & 21.6 & 41.3 & 83.2 & 54.5 & 72.4 & 16.4 & 82.4 & 58.4 \\
\midrule
SDFT (w/ ref.) & 26.4 & 39.4 & 83.1 & 55.0 & 72.1 & 15.3 & \textbf{83.0} & 58.0 \\
SDFT + SFT & 27.8 & 39.0 & 83.2 & 53.2 & 71.0 & 13.7 & 81.9 & 57.0 \\
SDFT + ACE & 28.9 & \textbf{\textit{41.6}} & 84.5 & \textbf{\textit{55.9}} & 72.1 & 16.5 & 82.5 & \textbf{\textit{58.9}} \\
\textbf{RoCo} & 28.9 & 40.8 & \textbf{\textit{85.4}} & \textbf{56.8} & \textbf{72.9} & \textbf{\textit{16.7}} & \textbf{\textit{82.9}} & \textbf{59.2} \\
\textbf{RoCo} + SFT & 29.5 & 40.1 & 82.2 & 55.1 & 71.9 & 15.0 & 81.2 & 57.6 \\
ACE only & \textbf{\textit{30.7}} & 38.9 & 70.7 & 52.8 & 70.6 & 13.9 & 80.8 & 54.6 \\
\textbf{RoCo-ACE} & \textbf{31.2} & \textbf{42.0} & \textbf{85.6} & 55.4 & \textbf{\textit{72.8}} & \textbf{16.8} & 82.7 & \textbf{59.2} \\
\bottomrule
\end{tabular}
}
\caption{Supplementary component ablations on Qwen3-VL-30B with full retention metrics. EVOKE measures injected-knowledge accuracy. Retention columns report held-out general multimodal and safety accuracy, and Ret. Avg. averages TreeBench, VStar, MathVision, MMStar, BabyVision, and MM-Safety. Higher values are better; bold and bold-italic mark the best and second-best scores in each column.}
\label{tab:qwen3vl30b_ablation_results}
\end{table*}


Table~\ref{tab:qwen3vl30b_results} keeps the benchmark columns and baseline groups consistent with the main Qwen3-VL-8B table to test whether the injection-retention trends persist at a larger model scale. The 30B results show the same qualitative pattern: direct fitting improves injected knowledge but sacrifices retention, constrained-update methods are more conservative, and \textbf{RoCo-ACE} gives the strongest injected-knowledge result while keeping retention close to or above the base model.

Table~\ref{tab:qwen3vl30b_ablation_results} reports component-level Qwen3-VL-30B ablations with every retention benchmark. The larger model follows the same trend as Qwen3-VL-8B: full-answer SFT controls improve EVOKE less efficiently and reduce retention, ACE improves target injection, and combining it with RoCo gives the best injected-knowledge accuracy among the compared variants while keeping the retention average close to the base model.


\section{Supplementary Results on InternVL3.5-8B}
\label{sec:internvl35_evoke_results}


\begin{table*}[!t]
\centering

\scriptsize
\setlength{\tabcolsep}{2.0pt}
\resizebox{2\columnwidth}{!}{
\begin{tabular}{lcccccccc}
\toprule
\multirow{2}{*}{\textbf{Variant}} &
\multirow{2}{*}{\textbf{EVOKE}} &
\multicolumn{7}{c}{\textbf{Retention Benchmarks}} \\
\cmidrule(lr){3-9}
& & \textbf{TreeBench} & \textbf{VStar} & \textbf{MathVision} & \textbf{MMStar} & \textbf{BabyVision} & \textbf{MM-Safety} & \textbf{Ret. Avg.} \\
\midrule
InternVL3.5-8B & 11.1 & \textbf{37.4} & 71.2 & \textbf{34.3} & \textbf{\textit{67.1}} & 11.1 & \textbf{75.8} & \textbf{49.5} \\
\midrule
\textbf{RoCo} & 17.8 & 36.3 & \textbf{71.9} & \textbf{\textit{34.0}} & 67.0 & \textbf{11.8} & 74.5 & 49.3 \\
ACE only & \textbf{\textit{19.2}} & 34.4 & 69.2 & 31.9 & 66.8 & 9.2 & 74.1 & 46.7 \\
\textbf{RoCo-ACE} & \textbf{20.9} & \textbf{\textit{36.7}} & \textbf{\textit{71.6}} & 34.0 & \textbf{67.5} & \textbf{\textit{11.6}} & \textbf{\textit{74.7}} & \textbf{\textit{49.3}} \\
\bottomrule
\end{tabular}
}
\caption{InternVL3.5-8B component comparison on EVOKE. EVOKE measures injected-knowledge accuracy. Retention columns report held-out multimodal and safety benchmark accuracy, and Ret. Avg. averages TreeBench, VStar, MathVision, MMStar, BabyVision, and MM-Safety. Higher values are better; bold and bold-italic mark the best and second-best scores in each column.}
\label{tab:internvl35_evoke_ablation_results}
\end{table*}

Table~\ref{tab:internvl35_evoke_ablation_results} reports an EVOKE component comparison on InternVL3.5-8B, testing whether the component-level trend of \textbf{RoCo-ACE} transfers beyond the Qwen3-VL model family.
The comparison follows the same protocol as the Qwen3-VL ablations: the base model is compared with \textbf{RoCo}, ACE-only training, and the full \textbf{RoCo-ACE} objective on EVOKE, while using the same six held-out retention benchmarks.

The results are consistent with the intended component behavior.
\textbf{RoCo} raises EVOKE Inj. Acc. from 11.1 to 17.8 while keeping Ret. Avg. nearly unchanged relative to the base model.
ACE-only training provides a stronger injection signal, reaching 19.2 EVOKE Inj. Acc., but lowers Ret. Avg. to 46.7, reflecting the cost of anchor correction without rollout-side distillation.
The full \textbf{RoCo-ACE} objective gives the best EVOKE Inj. Acc. of 20.9 while keeping Ret. Avg. at 49.3, close to the base model's 49.5.
This supports the same conclusion as the Qwen3-VL experiments: \textbf{RoCo} helps limit evaluated retention drift, while \textbf{ACE} supplies sparse correction for omitted authoritative anchors.




\section{Hyperparameter Sensitivity}
\label{sec:hyperparameter_sensitivity}

\begin{table*}[!t]
\centering

\scriptsize
\setlength{\tabcolsep}{2pt}
\renewcommand{\arraystretch}{0.92}
\resizebox{1.6\columnwidth}{!}{
\begin{tabular}{llccl}
\toprule
\textbf{Group} & \textbf{Setting} & \textbf{EVOKE} & \textbf{Ret. Avg.} & \textbf{Expected role} \\
\midrule
\multirow{3}{*}{$\lambda_{\mathrm{ACE}}$}
& $0.0$ & 23.7 & \textbf{56.6} & RoCo only; no anchored missing-fact correction \\
& $0.2$ & \textbf{\textit{27.6}} & \textbf{\textit{56.5}} & Default setting \\
& $1.0$ & \textbf{28.1} & 52.5 & Aggressive anchor fitting \\
\midrule
\multirow{3}{*}{$\beta_{\mathrm{miss}}$}
& $3.0$ & 25.1 & \textbf{56.8} & Moderate missing-span emphasis \\
& $5.0$ & \textbf{\textit{27.6}} & \textbf{\textit{56.5}} & Default setting \\
& $7.0$ & \textbf{28.2} & 54.1 & Aggressive missing-anchor emphasis \\
\midrule
\multirow{3}{*}{$\tau$}
& $0.1$ & 23.9 & \textbf{\textit{56.3}} & Mild contrast margin \\
& $0.2$ & \textbf{27.6} & \textbf{56.5} & Default setting \\
& $0.4$ & \textbf{\textit{25.2}} & 55.9 & Conservative contrast margin \\
\bottomrule
\end{tabular}
}
\caption{Hyperparameter sensitivity grid for \textbf{RoCo-ACE} on EVOKE. EVOKE measures injected-knowledge accuracy, and Ret. Avg. averages TreeBench, VStar, MathVision, MMStar, BabyVision, and MM-Safety. The RoCo-only and default rows reuse the corresponding results from Table~\ref{tab:main_baseline_results}; bold and bold-italic mark the best and second-best scores within each hyperparameter group.}
\label{tab:hyperparameter_ablation_results}
\end{table*}


Table~\ref{tab:hyperparameter_ablation_results} reports sensitivity analysis for the three hyperparameters that directly control the two components of \textbf{RoCo-ACE}. The ACE loss weight $\lambda_{\mathrm{ACE}}$ controls how strongly missing or weakly covered authoritative anchors are supervised. The missing-span multiplier $\beta_{\mathrm{miss}}$ controls how much more weight is assigned to anchors absent from the student rollout. The RoCo contrast margin $\tau$ controls how much the reference-conditioned teacher likelihood must exceed the base-teacher likelihood before a rollout token receives stronger distillation weight.

The $\lambda_{\mathrm{ACE}}$ rows show the role of anchored correction. Setting $\lambda_{\mathrm{ACE}}=0$ reduces the method to RoCo-only, which keeps Ret. Avg. high but lowers EVOKE accuracy to 23.7. The default $\lambda_{\mathrm{ACE}}=0.2$ improves EVOKE to 27.6 with almost no retention loss, while the more aggressive setting $\lambda_{\mathrm{ACE}}=1.0$ slightly increases EVOKE but drops Ret. Avg. to 52.5. This suggests that ACE should remain a targeted correction term rather than dominate the online-distillation objective.

The $\beta_{\mathrm{miss}}$ and $\tau$ rows show the same balance. Moderate missing-span emphasis under-corrects omitted facts, while overly large missing-span emphasis pressures retention. For the RoCo margin, $\tau=0.2$ gives the best injection-retention balance among the tested settings: a smaller margin is too permissive and a larger margin is too conservative. Overall, the default configuration is selected because it achieves strong injected-knowledge accuracy without turning training into full-reference imitation.


\section{Additional Discussion on Retention Scope and Reference Quality}
\label{sec:additional_discussion}

\textbf{RoCo-ACE} does not explicitly guarantee preservation of every general capability. Its preservation mechanism is indirect: it reduces irrelevant updates during injected-data training. This distinction matters because many regressions are not caused by learning a new fact itself, but by repeatedly training on long, narrow, stylistically similar answers. The method reduces that source of regression by making the update selective with respect to factual support from the reference.

The method also depends on reference quality. If the authoritative answer is noisy, overly broad, or stylistically biased, the RoCo contrast can amplify the wrong signal and ACE can reinforce incorrect spans. Span-level diagnostics are therefore important: the system should log which generated spans receive high contrast weights and which reference spans receive anchor correction. Such diagnostics make failure modes visible and help distinguish genuine factual injection from style imitation. The open-ended injection metric also relies on an LLM judge, so judge-agreement or human-audit results should be reported when the benchmark setting allows them.


\section{Benchmark Details}
\label{sec:benchmark_details}

\paragraph{Knowledge injection benchmarks.}
EVOKE \citep{jiang2025mmevoke} contains image-grounded entity and news updates with authoritative reference answers. This setting is deliberately different from rich-demonstration continual-learning settings studied by SDFT-style methods: the reference information is often short, sparse, and localized to a small number of factual spans, making uniform online distillation an incomplete injection signal. To test whether the method generalizes beyond time-sensitive news and entity updates, we additionally include two MLLM-CL benchmarks \citep{zhao2025mllmcl}: VP from the ability continual-learning suite and Sci from the domain continual-learning suite. MLLM-CL organizes ability data as non-IID tasks that introduce new visual capabilities, and domain data as IID tasks drawn from mainstream visual domains. In our setting, VP evaluates capability-oriented visual perception injection, while Sci evaluates science-domain knowledge injection.

We follow the original data splits released with each benchmark. EVOKE uses its provided entity/news knowledge-injection split with authoritative reference answers, and VP and Sci use the corresponding MLLM-CL training and test splits without additional resampling. All compared methods are trained and evaluated on the same split for each benchmark; no retention benchmark examples are mixed into the injection training set.

\paragraph{Retention benchmarks.}
TreeBench \citep{wang2026treebench} evaluates traceable visual grounded reasoning and evidence localization. VStar, following V*Bench \citep{wu2024vstar}, evaluates fine-grained visual search in high-resolution and cluttered scenes. MathVision \citep{wang2024mathvision} measures multimodal mathematical reasoning over visual problems. MMStar \citep{chen2024mmstar} evaluates vision-indispensable multimodal understanding with reduced language-only shortcuts and data leakage. BabyVision \citep{chen2026babyvision} probes core visual abilities that should be solved with minimal reliance on language priors. MM-SafetyBench \citep{liu2023mmsafetybench} evaluates multimodal safety behavior under image-conditioned unsafe instructions. Together, these benchmarks test whether knowledge injection preserves visual grounding, fine-grained perception, mathematical reasoning, general multimodal understanding, early visual primitives, and safety behavior.


\section{Evaluation Protocol}
\label{sec:evaluation_protocol}

We use accuracy as the primary metric. For open-ended injected-knowledge examples, we use Qwen3-30B-A3B-Instruct-2507 as the LLM judge \citep{qwen2025qwen3}. The judge compares each model response against the authoritative answer and returns a binary correctness label. It is instructed to focus on factual consistency with the reference, including entity identity, event identity, date, number, location, relation, and other task-specific facts, rather than surface wording. We report the resulting LLM-judge accuracy over the evaluation set.

For general capability benchmarks, we follow the benchmark-specific answer format and compute accuracy after answer normalization. For multiple-choice tasks, the predicted option is extracted from the model response. When the output is ambiguous or free-form, the same LLM-judge protocol is used to decide whether it matches the ground-truth answer. We report injected-knowledge accuracy together with general benchmark accuracy, since a method is only useful if it improves injected facts without causing substantial capability degradation.


\section{Model and Baseline Details}
\label{sec:model_baseline_details}

The main experiments use \textbf{Qwen3-VL-8B} as the base model because it exposes realistic retention failures while keeping the full baseline suite computationally tractable. We include \textbf{Qwen3-VL-30B} experiments in the supplementary material to check whether the observed trends persist at a larger scale. Unless otherwise stated, the vision tower is frozen, text and projector-side trainable components follow the corresponding baseline configuration, and \textbf{RoCo} uses one on-policy generation per prompt with a synchronized reference teacher.

\paragraph{Direct fitting baselines.}
LoRA \citep{hu2022lora} is the parameter-efficient fitting baseline, while SFT \citep{ouyang2022training} directly fits authoritative answers and follows the standard supervised instruction-tuning paradigm. These baselines test how much injected knowledge can be acquired through direct answer fitting and how much retention is lost under ordinary supervised updates.

\paragraph{Reference-conditioned self-distillation baseline.}
SDFT (w/ authoritative reference) \citep{shenfeld2026sdft} uses the same on-policy student rollout, synchronized teacher, and reference-conditioned teacher prompt as \textbf{RoCo}, but applies a uniform distillation loss over all valid rollout tokens. This is the main online-distillation reference point because every injected example provides an authoritative answer. A no-reference vanilla KD baseline is not included in the main comparison: without the authoritative reference, the teacher cannot observe the injected facts and the experiment mainly measures policy preservation rather than knowledge acquisition.

\paragraph{Constrained-update baselines.}
KORE \citep{jiang2025kore} represents knowledge-injection-specific constrained updating and is expected to be strong on target injection, but prior results also show that strong injection does not uniformly preserve every general capability dimension. Following the KORE protocol, we sample auxiliary general examples from LLaVA-OneVision \citep{li2024llavaonevision} to estimate the general representation space used by its retention constraint; these auxiliary examples are not used by our method or by other baselines. KeepLoRA \citep{luo2026keeplora}, MoELoRA \citep{luo2024moelora}, and SEFE \citep{chen2025sefe} represent parameter-space constrained adaptation methods that reduce forgetting through residual-gradient updates, expertized LoRA modules, or forgetting-aware constraints. Among them, KeepLoRA \citep{luo2026keeplora} is a retention-oriented baseline that can preserve general behavior more conservatively but may provide weaker injection when the new knowledge is sparse.

\paragraph{Our variants.}
The main variants are \textbf{RoCo} and \textbf{RoCo-ACE}. The former isolates rollout-conditioned contrastive online distillation over student rollouts, while the latter combines it with missing-span anchored supervision. Additional component controls, including SDFT+SFT, SDFT+ACE, \textbf{RoCo}+SFT, and ACE only, are reported in the ablation tables.

\paragraph{Implementation form of baselines.}
For Qwen3-VL-8B, SFT, SDFT, and \textbf{RoCo} variants use full language-side updating with the vision tower and multimodal aligner frozen. LoRA, KeepLoRA, MoELoRA, SEFE, and KORE are implemented as adapter-based or constrained-update baselines following their method definitions, using the same rank, target-module, and training-budget settings unless the method requires its own constraint or two-stage procedure. For Qwen3-VL-30B, the non-adapter methods likewise use full language-side updating with frozen vision and aligner modules, while adapter-defined methods remain adapter-based. This keeps the comparison aligned with the intended update mechanism of each baseline rather than converting all methods into a single adapter form.


\section{Training Configuration and Hyperparameters}
\label{sec:training_config}

Unless otherwise stated, all primary experiments are run for five epochs with learning rate $1\times10^{-5}$, warmup ratio $0.05$, bfloat16 precision, FlashAttention, DeepSpeed ZeRO-3, per-device batch size $1$, and gradient accumulation $8$. We set the maximum text sequence length to $1536$ and the maximum generated completion length to $1024$. For image inputs, we set the minimum and maximum pixel budgets to $65{,}536$ and $3{,}379{,}200$, corresponding to an image-token range of $64$ to $3300$. The vision tower and multimodal aligner are frozen. Full-model updating is used for SFT, SDFT, and \textbf{RoCo} variants, while adapter-based baselines use LoRA-style updates with rank $16$, alpha $32$, dropout $0.05$, and all linear layers as target modules.

For all online distillation variants, the student samples one on-policy completion per prompt (\texttt{num\_generations}=1) with temperature $1.0$, top-$p=1.0$, and one return sequence. The teacher uses the same base model architecture and the same Qwen3-VL chat template as the student. The student prompt contains only the target user question and the optional image. The reference-conditioned teacher prompt replaces the user message with a reference-augmented instruction of the form: target question, authoritative reference answer, and an instruction to generate a new answer that is factually consistent with the reference. The base teacher prompt uses the original student input without the authoritative reference. The teacher never generates the training completion in our main setting; it only scores the student's rollout under these two prompt contexts. The teacher is synchronized with the student by an exponential-moving-average update after every optimization step, so the teacher tracks the current policy while remaining a stable scoring model.

\paragraph{Teacher prompt template.}
For English reference-conditioned teacher scoring, we use the following template, where \texttt{\{question\}} is the original user query and \texttt{\{gold\_response\}} is the authoritative reference answer:
\begin{center}
\setlength{\fboxsep}{4pt}
\fbox{\begin{minipage}{0.92\columnwidth}
\footnotesize\ttfamily\raggedright
Target Question:\par
\{question\}\par\medskip
Response to the target question:\par
\{gold\_response\}\par\medskip
The assistant answer above is the authoritative reference for this training example.\par
Generate a new answer that stays fully consistent with the same entity/event identity and factual claims.\par
Use the image only to add visible grounding details and natural wording.\par
Do not contradict, negate, or correct the reference answer.\par
Do not claim that the image is unrelated, blank
\end{minipage}}
\end{center}

For \textbf{RoCo}, we set the distillation loss weight $\lambda_{\mathrm{RoCo}}=1.0$, the floor token weight $w_0=1.0$, the reference-contrast scale $\beta_\Delta=1.0$, and the contrast margin $\tau=0.2$. Dense token weighting, span aggregation, and weight normalization are enabled. We do not use top-$m$ truncation or any auxiliary gating for rollout-conditioned contrastive tokens. The synchronized teacher is updated after every optimization step with exponential moving average decay $0.999$, corresponding to an effective update coefficient of $0.001$ from the current student.

For Anchored Cross-Entropy, we set the objective weight $\lambda_{\mathrm{ACE}}=0.2$ unless otherwise stated. The base weight inside extracted anchor spans is $u_0=1.0$, and the missing-span weight is $\beta_{\mathrm{miss}}=5.0$; reference tokens outside extracted factual anchors are not included in the ACE loss. Missing-span weighting and span aggregation are enabled for ACE variants. For SDFT (w/ ref.), we keep the same student rollout and reference-conditioned teacher-scoring pipeline as \textbf{RoCo}, but replace the rollout-conditioned contrastive weights with uniform weights and set $\lambda_{\mathrm{ACE}}=0$. For hybrid SDFT+SFT and \textbf{RoCo}+SFT variants, the supervised loss weight is set to $1.0$ over the full reference answer.

\begin{table}[!t]
\centering
\scriptsize
\setlength{\tabcolsep}{4pt}
\renewcommand{\arraystretch}{1.05}
\resizebox{\columnwidth}{!}{
\begin{tabular}{lcc}
\toprule
\textbf{Method} & \textbf{Training time (h)} & \textbf{Inference latency} \\
\midrule
LoRA & $\sim$5.0 & $\sim$1.0 s/query \\
SFT & $\sim$10.0 & $\sim$1.0 s/query \\
SDFT (w/ ref.) & $\sim$20.0 & $\sim$1.0 s/query \\
KORE & $\sim$9.0 & $\sim$1.0 s/query \\
KeepLoRA & $\sim$6.0 & $\sim$1.0 s/query \\
MoELoRA & $\sim$5.0 & $\sim$1.0 s/query \\
SEFE & $\sim$9.0 & $\sim$1.0 s/query \\
\textbf{RoCo-ACE} & $\sim$22.0 & $\sim$1.0 s/query \\
\bottomrule
\end{tabular}
}
\caption{Estimated runtime comparison on the EVOKE setting with Qwen3-VL-8B, three training epochs, and 16 PPU accelerators. Training time is reported as wall-clock hours from representative runs. Inference latency is similar across methods because the deployed model sizes are comparable; we observe roughly one second per query under the same inference setup.}
\label{tab:runtime_comparison}
\end{table}


\paragraph{Runtime cost.}
Table~\ref{tab:runtime_comparison} reports approximate wall-clock training and inference cost on EVOKE with Qwen3-VL-8B, three training epochs, and 16 PPU accelerators. \textbf{RoCo-ACE} is slower than direct fitting and adapter-only baselines because it performs online rollout scoring with reference-conditioned and base-teacher contexts, but its cost remains close to SDFT since both use the same online-distillation backbone. At inference time, all methods use model weights of comparable scale, so their per-query latency remains approximately the same under the same decoding setup.


\end{document}